\PassOptionsToPackage{unicode}{hyperref}
\PassOptionsToPackage{hyphens}{url}
\PassOptionsToPackage{dvipsnames,svgnames,x11names}{xcolor}
\documentclass[
]{article}
\usepackage{amsmath,amssymb}
\usepackage{iftex}
\ifPDFTeX
  \usepackage[T1]{fontenc}
  \usepackage[utf8]{inputenc}
  \usepackage{textcomp} 
\else 
  \usepackage{unicode-math} 
  \defaultfontfeatures{Scale=MatchLowercase}
  \defaultfontfeatures[\rmfamily]{Ligatures=TeX,Scale=1}
\fi
\usepackage{lmodern}
\ifPDFTeX\else
\fi
\IfFileExists{upquote.sty}{\usepackage{upquote}}{}
\IfFileExists{microtype.sty}{
  \usepackage[]{microtype}
  \UseMicrotypeSet[protrusion]{basicmath} 
}{}
\makeatletter
\@ifundefined{KOMAClassName}{
  \IfFileExists{parskip.sty}{%
    \usepackage{parskip}
  }{
    \setlength{\parindent}{0pt}
    \setlength{\parskip}{6pt plus 2pt minus 1pt}}
}{
  \KOMAoptions{parskip=half}}
\makeatother
\usepackage{xcolor}
\usepackage[margin=1in]{geometry}
\usepackage{color}
\usepackage{fancyvrb}

\DefineVerbatimEnvironment{Highlighting}{Verbatim}{commandchars=\\\{\}}
\usepackage{framed}
\definecolor{shadecolor}{RGB}{248,248,248}
\newenvironment{Shaded}{\begin{snugshade}}{\end{snugshade}}

\newcommand{\BuiltInTok}[1]{#1}

\newcommand{\CommentTok}[1]{\textcolor[rgb]{0.56,0.35,0.01}{\textit{#1}}}

\newcommand{\DecValTok}[1]{\textcolor[rgb]{0.00,0.00,0.81}{#1}}

\newcommand{\FloatTok}[1]{\textcolor[rgb]{0.00,0.00,0.81}{#1}}

\newcommand{\ImportTok}[1]{#1}

\newcommand{\NormalTok}[1]{#1}
\newcommand{\OperatorTok}[1]{\textcolor[rgb]{0.81,0.36,0.00}{\textbf{#1}}}

\newcommand{\StringTok}[1]{\textcolor[rgb]{0.31,0.60,0.02}{#1}}
\newcommand{\VariableTok}[1]{\textcolor[rgb]{0.00,0.00,0.00}{#1}}

\usepackage{graphicx}
\makeatletter
\newsavebox\pandoc@box
\newcommand*\pandocbounded[1]{
  \sbox\pandoc@box{#1}%
  \Gscale@div\@tempa{\textheight}{\dimexpr\ht\pandoc@box+\dp\pandoc@box\relax}%
  \Gscale@div\@tempb{\linewidth}{\wd\pandoc@box}%
  \ifdim\@tempb\p@<\@tempa\p@\let\@tempa\@tempb\fi
  \ifdim\@tempa\p@<\p@\scalebox{\@tempa}{\usebox\pandoc@box}%
  \else\usebox{\pandoc@box}%
  \fi%
}
\def\fps@figure{htbp}
\makeatother
\NewDocumentCommand\citeproctext{}{}
\NewDocumentCommand\citeproc{mm}{%
  \begingroup\def\citeproctext{#2}\cite{#1}\endgroup}
\makeatletter
 \let\@cite@ofmt\@firstofone
 \def\@biblabel#1{}
 \def\@cite#1#2{{#1\if@tempswa , #2\fi}}
\makeatother
\newlength{\cslhangindent}
\newlength{\csllabelwidth}
\newenvironment{CSLReferences}[2] 
 {\begin{list}{}{%
  \setlength{\itemindent}{0pt}
  \setlength{\leftmargin}{0pt}
  \setlength{\parsep}{0pt}
  \ifodd #1
   \setlength{\leftmargin}{\cslhangindent}
   \setlength{\itemindent}{-1\cslhangindent}
  \fi
  \setlength{\itemsep}{#2\baselineskip}}}
 {\end{list}}
\usepackage{calc}

\usepackage{booktabs}
\usepackage{caption}
\usepackage{longtable}
\usepackage{colortbl}
\usepackage{array}
\usepackage{anyfontsize}
\usepackage{multirow}
\usepackage{bookmark}
\IfFileExists{xurl.sty}{\usepackage{xurl}}{} 
\hypersetup{
  pdftitle={Non-Negative Stiefel Approximating Flow: Orthogonalish Matrix Optimization for Interpretable Embeddings},
  colorlinks=true,
  linkcolor={blue},
  filecolor={Maroon},
  citecolor={Blue},
  urlcolor={blue},
  pdfcreator={LaTeX via pandoc}}

\title{Non-Negative Stiefel Approximating Flow: Orthogonalish Matrix
Optimization for Interpretable Embeddings}
\author{Brian B. Avants, Nicholas J. Tustison, and James R. Stone\\
Department of Radiology and Medical Imaging\\
University of Virginia, Charlottesville, VA 22903}
\date{October 11, 2025}

\begin{document}
\maketitle

\section*{Abstract}\label{abstract}
\addcontentsline{toc}{section}{Abstract}

Interpretable representation learning is a central challenge in modern
machine learning, particularly in high-dimensional settings such as
neuroimaging, genomics, and text analysis. Current methods often
struggle to balance the competing demands of interpretability and model
flexibility, limiting their effectiveness in extracting meaningful
insights from complex data. We introduce Non-negative Stiefel
Approximating Flow (NSA-Flow), a general-purpose matrix estimation
framework that unifies ideas from sparse matrix factorization,
orthogonalization, and constrained manifold learning. NSA-Flow enforces
structured sparsity through a continuous balance between reconstruction
fidelity and column-wise decorrelation, parameterized by a single
tunable weight. The method operates as a smooth flow near the Stiefel
manifold with proximal updates for non-negativity and adaptive gradient
control, yielding representations that are simultaneously sparse,
stable, and interpretable. Unlike classical regularization schemes,
NSA-Flow provides an intuitive geometric mechanism for manipulating
sparsity at the level of global structure while simplifying latent
features. We demonstrate that the NSA-Flow objective can be optimized
smoothly and integrates seamlessly with existing pipelines for
dimensionality reduction while improving interpretability and
generalization in both simulated and real biomedical data. Empirical
validation on the Golub leukemia dataset and in Alzheimer's disease
demonstrate that the NSA-Flow constraints can maintain or improve
performance over related methods with little additional methodological
effort. NSA-Flow offers a scalable, general-purpose tool for
interpretable ML, applicable across data science domains.

\section{Introduction}\label{introduction}

Modern machine learning increasingly faces the challenge of extracting
interpretable structure from high-dimensional, correlated data. In
domains such as neuroscience, genomics, or natural language processing,
data matrices often encode overlapping sources of variation: voxels
representing distributed brain activity, genes co-expressed across
pathways or words co-occurring across topics. These correlations hinder
modeling, making it difficult to disentangle meaningful latent factors
arising from complex phenomena such as gene expression profiles in
bioinformatics (\citeproc{ref-golub1999molecular}{Golub et al. 1999}),
term-document frequencies in topic modeling
(\citeproc{ref-blei2003latent}{Blei, Ng, and Jordan 2003}), multi-view
biological measurements in integrative omics
(\citeproc{ref-strazar2016orthogonal}{Stražar and Žitnik 2016}), and
user-item interactions in recommender systems
(\citeproc{ref-koren2009matrix}{Koren, Bell, and Volinsky 2009}).

Classical dimensionality reduction techniques like principal component
analysis (PCA) and its sparse variants seek low-rank approximations with
interpretable bases (\citeproc{ref-zou2006sparse}{Zou, Hastie, and
Tibshirani 2006}), while sparse canonical correlation analysis (CCA)
extends this to multi-view correlations in biological data
(\citeproc{ref-witten2011}{Witten, Tibshirani, and Hastie 2009}).
However, enforcing sparsity and decorrelation remains challenging:
traditional methods may over-regularize or lack intuitive controls for
partial constraints. Non-negative matrix factorization (NMF)
(\citeproc{ref-lee2001nmf}{Daniel D. Lee and Seung 2001b}) offers
parts-based, additive representations aligned with domain constraints,
but suffers from rotational ambiguity, yielding entangled factors
(\citeproc{ref-ding2006orthogonal}{Ding et al. 2006}). Orthogonal
variants improve sparsity and identifiability by aligning factors with
disjoint structures, with applications in sparse PCA
(\citeproc{ref-zou2006sparse}{Zou, Hastie, and Tibshirani 2006}), sparse
CCA (\citeproc{ref-witten2011}{Witten, Tibshirani, and Hastie 2009}),
and interpretable neural networks (\citeproc{ref-henaff2011deep}{Henaff
et al. 2011}). Yet, strict orthogonality often sacrifices fidelity,
especially in noisy or heterogeneous data and motivates the need for
partial decorrelation models.

Soft orthogonalization methods address this by penalizing deviations
from orthonormality. Examples include Disentangled Orthogonality
Regularization (DOR), which separates Gram matrix components for
convolutional kernels (\citeproc{ref-wu2023towards}{Wu 2023}); Group
Orthogonalization Regularization (GOR), which applies intra-group
penalties for vision tasks (\citeproc{ref-kurtz2023group}{Kurtz, Bar,
and Giryes 2023}); \(\lambda\)-Orthogonality Regularization, which
introduces thresholded penalties for representation learning
(\citeproc{ref-ricci2025orthogonality}{Ricci et al. 2025}); and simpler
approaches like Spectral Restricted Isometry Property (SRIP)
(\citeproc{ref-goessmann2020restricted}{Goeßmann 2020}) and Frobenius
norm penalties for neural stability (\citeproc{ref-guo2019frobenius}{Guo
2019}). However, these methods are typically embedded in neural training
pipelines and do not enforce non-negativity, limiting their
applicability to domains where measurement units should be preserved.
Advanced ONMF variants, such as variational Bayesian approaches
(\citeproc{ref-rahiche2022variational}{Rahiche et al. 2022}), unilateral
factorization (\citeproc{ref-li2023unilateral}{Li, Zhang, and Zhang
2023}), and deep autoencoder frameworks
(\citeproc{ref-yang2021orthogonal}{Yang and Xu 2021}), improve
robustness but enforce strict orthogonality or require full
decomposition, reducing flexibility for one-sided refinement.

In contrast to these soft regularization techniques, Riemannian
optimization approaches have been explored primarily for enforcing
strict orthogonality constraints on manifolds such as the Stiefel
manifold, where the feasible set is equipped with a differential
structure for gradient-based updates. For instance, Nonlinear Riemannian
Conjugate Gradient (NRCG) methods optimize orthogonal NMF by projecting
gradients onto the tangent space and using retractions like QR
decomposition to maintain exact orthonormality, while handling
non-negativity through coordinate descent on the complementary factor
(\citeproc{ref-Zhang2016EfficientON}{Zhang et al. 2016}). This ensures
convergence to critical points with near-perfect orthogonality but
incurs higher computational costs compared to soft penalties, and it
typically requires full enforcement rather than flexible deviations.
Hybrid methods like Feedback Gradient Descent (FGD) attempt to bridge
this gap by approximating manifold dynamics in Euclidean space with
feedback terms to achieve stable near-orthogonality efficiently,
outperforming traditional Riemannian methods in DNN training speed while
rivaling soft constraints in overhead
(\citeproc{ref-Bu2022FeedbackGD}{Bu and Chang 2022}). However, adapting
such Riemannian-inspired techniques to incorporate soft orthogonality
penalties alongside non-negativity remains a less explored avenue.

To address these gaps, we propose \textbf{Non-negative Stiefel
Approximating Flow (NSA-Flow)}, a variational optimization algorithm
that approximates a target ( \(X_0\) ) with a non-negative matrix (
\(Y \in \mathbb{R}^{p \times k}_{\geq 0}\) ). NSA-Flow balances
fidelity, column orthogonality, and non-negativity through a single
tunable parameter ( \(w \in [0,1]\) ) influencing proximity to the
Stiefel manifold. The simple parameterization of these constraints means
that NSA-Flow allows practitioners to directly encode the desired level
of sparsity/decorrelation (which are closely related in this framework)
in their embeddings, without the need for complex regularization schemes
or full orthogonality constraints. As such, NSA-Flow employs global soft
orthogonality constraints to promote disjoint support across columns and
foster interpretable bases. Formulated to stay near the Stiefel manifold
(\citeproc{ref-edelman1998geometry}{Edelman, Arias, and Smith 1998}), it
is inspired by Riemannian gradient descent
(\citeproc{ref-absil2008optimization}{Absil, Mahony, and Sepulchre
2008}) with flexible retractions (purely Euclidean, polar retraction or
a novel soft interpolation between these). Non-negativity is ensured via
proximal projections (\citeproc{ref-parikh2014proximal}{Parikh and Boyd
2014}), maintaining descent stability and constraint satisfaction.
Conceptually, NSA-Flow functions as a soft projection operator that can
be inserted into any existing machine learning system to improve
interpretability---whether as a regularization layer in a neural network
(\citeproc{ref-henaff2011deep}{Henaff et al. 2011};
\citeproc{ref-ricci2025orthogonality}{Ricci et al. 2025}), a refinement
step in factor models (\citeproc{ref-li2023unilateral}{Li, Zhang, and
Zhang 2023}; \citeproc{ref-rahiche2022variational}{Rahiche et al.
2022}), or a sparsity-enforcing module in linear embeddings
(\citeproc{ref-guo2019frobenius}{Guo 2019};
\citeproc{ref-goessmann2020restricted}{Goeßmann 2020}). Unlike purely
regularization-based methods (\citeproc{ref-wu2023towards}{Wu 2023};
\citeproc{ref-kurtz2023group}{Kurtz, Bar, and Giryes 2023};
\citeproc{ref-ricci2025orthogonality}{Ricci et al. 2025}), NSA-Flow may
also operate as a one-sided projection operator, preserving input
structure while enabling controlled decorrelation.

Our contributions are:

\begin{enumerate}
\def\labelenumi{\arabic{enumi}.}
\item
  A general framework for constrained matrix approximation,
  parameterized by ( \(w\) ) to intuitively control sparsity and
  orthogonality for interpretable ML.
\item
  Rigorous empirical validation that demonstrates good convergence
  properties and reliable benchmark performance compared to baselines.
\item
  Broad applications, including enhanced disease classification on the
  Golub leukemia dataset, non-negative sparse PCA for biological
  integration and interpretable brain network discovery---showcasing
  NSA-Flow's versatility across ML domains.
\item
  An \href{https://github.com/stnava/nsa-flow}{open-source
  implementation} in \texttt{pytorch} facilitating easy integration into
  existing workflows for researchers and practitioners. NSA-flow is also
  \texttt{pip} installable via the \texttt{nsa-flow} package and wrapped
  in \texttt{R} via \texttt{ANTsR}.
\end{enumerate}

The paper is organized as follows: Section 2 derives the formulation and
algorithm; Section 3 details the experimental results; Section 4
discusses limitations and future work; Section 5 gives an overview of
software resources.

\section{Methods}\label{methods}

We consider the problem of finding a matrix
\(Y \in \mathbb{R}^{p \times k}\) that optimally approximates a target
matrix \(X_0 \in \mathbb{R}^{p \times k}\) while satisfying column
orthogonality and, optionally, non-negativity. This general formulation
is central to a wide range of problems in machine learning and signal
processing, including orthogonal dictionary learning, Independent
Component Analysis (ICA), and the orthogonal Procrustes problem
(\citeproc{ref-lee2001algorithms}{Daniel D. Lee and Seung 2001a};
\citeproc{ref-hyvarinen2000independent}{Hyvärinen and Oja 2000};
\citeproc{ref-schonemann1966generalized}{Schönemann 1966}). The NSA-Flow
optimization problem is defined by the minimization of a composite
energy function \(E(Y)\):

\[
\min_{Y \in \mathbb{R}^{p \times k}, Y \ge 0}
E(Y)
= (1 - w) \, L_{fid}(Y, X_0)
+ w \, L_{orth}(Y),
\label{eq:objective}
\]

where \(w \in [0, 1]\) is a hyperparameter that balances the fidelity
loss \(L_{fid}\) against the orthogonality loss \(L_{orth}\). For
numerical stability, the loss terms are internally re-weighted based on
their initial magnitudes, but we omit these scaling factors for
notational clarity. The fidelity term is the standard squared Frobenius
norm distance, \(L_{fid}(Y, X_0) = \frac{1}{2} \| Y - X_0 \|_F^2\). One
choice for orthogonality loss is: \[
L_{orth}(Y) = \frac{1}{2} \| Y^\top Y - I_k \|_F^2.
\] This penalty is zero if and only if \(Y\) belongs to the Stiefel
manifold
\(St(p, k) = \{ Y \in \mathbb{R}^{p \times k} : Y^\top Y = I_k \}\) and
grows quadratically with the orthogonality defect. The Euclidean
gradient of this objective is: \[
\nabla_Y E(Y) = (1 - w)(Y - X_0) + w \, Y ( Y^\top Y - I_k ).
\]

\subsection{The Optimization Challenge and the NSA-Flow
Approach}\label{the-optimization-challenge-and-the-nsa-flow-approach}

A standard Euclidean gradient descent step,
\(Y \leftarrow Y - \eta \nabla_Y E(Y)\), is ill-suited for this problem
as it does not respect the orthogonality constraint. The conventional
solution is to employ Riemannian optimization methods, which involve
projecting the gradient onto the tangent space of the Stiefel manifold
and then using a \textbf{retraction} to pull the updated iterate back
onto the manifold (\citeproc{ref-absil2008optimization}{Absil, Mahony,
and Sepulchre 2008}; \citeproc{ref-boumal2023intro}{Boumal 2023}). While
theoretically sound, full retraction steps can be computationally
expensive (requiring an SVD or matrix square root inverse) and can
sometimes hinder convergence on the fidelity term by abruptly correcting
the geometry of the iterate.

NSA-Flow introduces a \textbf{soft-retraction flow} that elegantly
circumvents these issues. Instead of enforcing a hard constraint at
every step, it defines an update rule that simultaneously descends on
the energy landscape and continuously ``pulls'' the iterates towards the
Stiefel manifold. This is achieved by directly linking the update rule's
geometry to the objective function's weight parameter \(w\). Figure 1
illustrates the conceptual framework of NSA-Flow, showing how the
iterates evolve under different \(w\) settings.

\subsubsection{The Soft-Retraction Method: A Geometrically-Aware
Update}\label{the-soft-retraction-method-a-geometrically-aware-update}

Let \(\widetilde{Y}^{(t+1)}\) be the iterate after a standard Euclidean
gradient step from \(Y^{(t)}\): \[
\widetilde{Y}^{(t+1)} = Y^{(t)} - \eta \, \nabla_Y E(Y^{(t)}).
\] Let
\(Q^{(t+1)} = \mathrm{Retract}(\widetilde{Y}^{(t+1)}) = \widetilde{Y}^{(t+1)} (\widetilde{Y}^{(t+1)\top} \widetilde{Y}^{(t+1)})^{-1/2}\)
be the polar retraction of \(\widetilde{Y}^{(t+1)}\), which is the
closest point to \(\widetilde{Y}^{(t+1)}\) on the Stiefel manifold in
the Frobenius norm (\citeproc{ref-edelman1998geometry}{Edelman, Arias,
and Smith 1998}). The NSA-Flow update is a convex combination of the
gradient step and its retraction, where the interpolation parameter is
the objective weight \(w\) itself: \[
Y^{(t+1)} = (1 - w) \, \widetilde{Y}^{(t+1)} + w \, Q^{(t+1)}.
\] This design choice creates a powerful, self-consistent algorithm:

\begin{itemize}
\item
  When \(w\) is small (fidelity is prioritized), the update is mostly a
  standard Euclidean gradient step.
\item
  When \(w\) is large (orthogonality is prioritized), the update is
  strongly pulled towards the manifold via the polar retraction.
\end{itemize}

This method can be viewed as an instance of an \textbf{averaged
operator} scheme, which is known to exhibit stable and smooth
convergence properties (\citeproc{ref-bauschke2017convex}{Bauschke and
Combettes 2017}). It provides a computationally efficient and
geometrically intuitive alternative to full Riemannian optimization,
similar in spirit to other fast, retraction-free, or approximate
manifold methods (\citeproc{ref-vary2024optimization}{Vary et al. 2024};
\citeproc{ref-ablin2022fast}{Ablin and Peyré 2022}).

\subsubsection{Geometric Stability and
Convergence}\label{geometric-stability-and-convergence}

The stability of the soft-retraction flow is guaranteed by its
contractive nature with respect to the constraint set.

\textbf{Proposition:} Let \(\tilde{Y} \in \mathbb{R}^{p \times k}\),
\(Q = \mathrm{Retract}(\tilde{Y})\), and
\(Y_{\text{new}} = (1-w)\tilde{Y} + w Q\) for \(w \in [0, 1]\). The
Frobenius distance of the new iterate to the Stiefel manifold is
strictly reduced for any \(w > 0\): \[
\| Y_{\text{new}} - Q \|_F = (1 - w) \| \tilde{Y} - Q \|_F.
\] \emph{Proof.} The proof follows directly from the linearity of the
norm:
\(\| Y_{\text{new}} - Q \|_F = \| ((1-w)\tilde{Y} + w Q) - Q \|_F = \| (1-w)(\tilde{Y} - Q) \|_F = (1 - w) \| \tilde{Y} - Q \|_F.\)
Since \(Q\) is the projection of \(\tilde{Y}\) onto the manifold, this
proposition shows that each soft-retraction step reduces the iterate's
distance to the feasible set. This property, fundamental to proximal
point and averaged operator algorithms, ensures that the iterates are
progressively and smoothly drawn towards the manifold, preventing
divergence and promoting stable convergence
(\citeproc{ref-parikh2014proximal}{Parikh and Boyd 2014};
\citeproc{ref-combettes2011proximal}{Combettes and Pesquet 2011}).

\subsubsection{Scale-Invariant Orthogonality
Penalty}\label{scale-invariant-orthogonality-penalty}

To enhance robustness, NSA-Flow's default setting uses a
\textbf{scale-invariant orthogonality defect}. The standard penalty
\(\| Y^\top Y - I \|_F^2\) is sensitive to the norm of \(Y\), as scaling
\(Y \to cY\) scales the penalty by \(c^4\). This can lead to poorly
conditioned optimization problems where the learning rate must be
carefully tuned. Following the principles in
(\citeproc{ref-wen2013feasible}{Wen, Yin, and Zhang 2013}), we use a
normalized penalty that is invariant to the global scale of \(Y\): \[
L_{orth, inv}(Y) = \frac{\| Y^\top Y - \text{diag}(\text{diag}(Y^\top Y)) \|_F^2}{\|Y\|_F^4}.
\] This objective purely measures the cosine of the angles between
columns, decoupling the orthogonality constraint from the magnitude of
the column vectors. This results in a better-conditioned optimization
landscape and more consistent convergence behavior.

\subsubsection{Relationship to Alternative Manifold Optimization
Methods}\label{relationship-to-alternative-manifold-optimization-methods}

\begin{itemize}
\item
  \textbf{Cayley Transform:} An alternative for preserving orthogonality
  is the \textbf{Cayley transform}, which defines an exact
  retraction-free update. For a skew-symmetric matrix
  \(A = \text{grad}E(Y) Y^\top - Y (\text{grad}E(Y))^\top\), the update
  \(Y^{(t+1)} = ( I - \frac{\eta}{2} A )^{-1} ( I + \frac{\eta}{2} A ) Y^{(t)}\)
  exactly preserves orthogonality
  (\citeproc{ref-gao2019parallelizing}{Gao et al. 2019}). However, it
  requires solving a \(p \times p\) linear system, making it
  computationally prohibitive for large \(p\).
\item
  \textbf{Riemannian Optimization Frameworks:} Standard toolboxes like
  \texttt{Manopt} (\citeproc{ref-boumal2014manopt}{Boumal et al. 2014})
  implement sophisticated algorithms like Riemannian trust-region and
  conjugate gradient methods. NSA-Flow's soft-retraction can be seen as
  a computationally efficient alternative that can be easily integrated
  into existing, traditional ML or deep learning pipelines and that
  allows for flexible constraint satisfaction.
\end{itemize}

\subsubsection{Computational Complexity}\label{computational-complexity}

\begin{table}[t]
\caption{\label{tab:complexity_table}Computational Complexity of NSA-Flow and Related Methods} 
\fontsize{6.0pt}{7.0pt}\selectfont
\begin{tabular*}{\linewidth}{@{\extracolsep{\fill}}lllll}
\toprule
Method & Dominant Operation & Complexity (\$p \textbackslash{}ge k\$) & Orthogonality & Notes \\ 
\midrule\addlinespace[2.5pt]
Euclidean GD & Gradient Computation & ( O(pk\^{}2) ) & ✗ (Unstable) & Fails to enforce constraints. \\ 
Full Polar Retraction & Gradient + SVD / Polar & ( O(pk\^{}2) ) & ✓ (Exact) & Costly, non-smooth updates. \\ 
\textbf{NSA-Flow (ours)} & Gradient + Polar + Interp. & ( O(pk\^{}2) ) & ≈ (Controlled) & Smoother, faster, practical. \\ 
Cayley Transform & Linear Solve ((p \textbackslash{}times p)) & ( O(p\^{}3) ) or ( O(pk\^{}2) ) (w/ low rank) & ✓ (Exact) & Prohibitive for large (p). \\ 
\bottomrule
\end{tabular*}
\end{table}

For tall-skinny matrices (\(p \gg k\)), the asymptotic cost is dominated
by \(O(pk^2)\) matrix multiplications. The primary advantage of the
soft-retraction flow is not in its asymptotic complexity but in its
superior convergence dynamics, offering a stable and smooth optimization
trajectory that effectively balances multiple objectives at a reduced
computational cost per iteration. Practically speaking, as will be shown
in the Results section, NSA-Flow provides an intuitive ``knob'' that
allows users to impose orthogonality, non-negativity and sparseness via
a consistent and intuitive framework. Table 1 summarizes these
comparisons.

\begin{figure}

{\centering \includegraphics{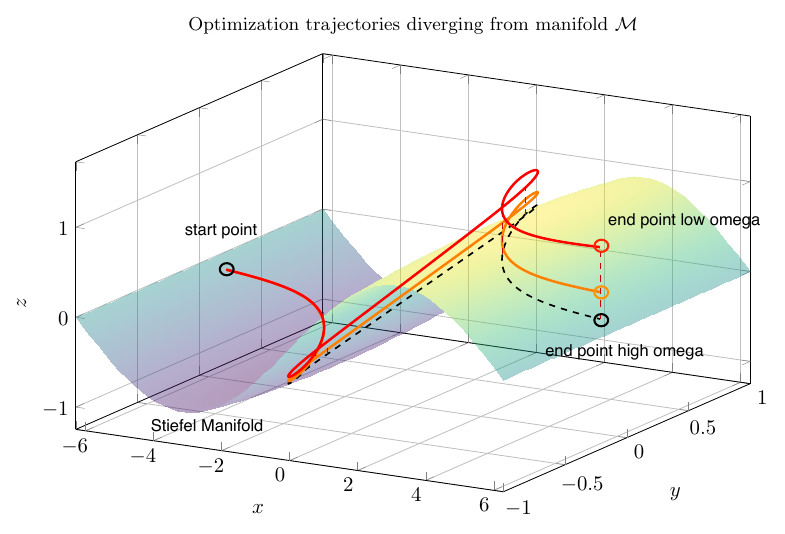} 

}

\caption{Illustration of the NSA-Flow optimization as a function of $\omega$.  The colored manifold is a conceptual representation of the Stiefel manifold, with the curves representing optimization paths for evolving $Y$.  When $\omega$ is small, the retraction is mild, allowing more deviation from orthonormality; when $\omega$ is large, the retraction strongly enforces orthonormality, pulling $Y$ closer to the manifold.}\label{fig:pollen}
\end{figure}

\subsubsection{Implementation}\label{implementation}

The NSA-Flow algorithm is implemented in \texttt{pytorch} (and wrapped
in \texttt{R}) as a modular, numerically stable framework for optimizing
non-negative matrices under orthogonality constraints. The main
function, \texttt{nsa\_flow\_orth}, is supported by helper functions for
matrix operations, gradient computations, retractions, and optimization.
Key design principles include robustness to numerical issues,
flexibility in retraction choices, and comprehensive diagnostics for
monitoring convergence (\citeproc{ref-absil2008optimization}{Absil,
Mahony, and Sepulchre 2008}).

The main function accepts an initial matrix \(Y_0\), an optional target
\(X_0\), and parameters for the orthogonality weight \(w\), retraction
type, maximum iterations, tolerance, and optimizer type (defaulting to
Averaged Stochastic Gradient Descent (ASGD)). If no \(X_0\) is provided,
\(Y_0\) becomes the target. The algorithm initializes scaling factors
for fidelity and orthogonality terms based on initial estimates of the
optimization landscape, ensuring balanced contributions across matrix
sizes and data content.

Each iteration computes the Euclidean gradients for fidelity and
orthogonality and then projects them toward the Stiefel manifold's
tangent space (\citeproc{ref-edelman1998geometry}{Edelman, Arias, and
Smith 1998}), and performs a descent step using an adaptive learning
rate. Retraction (polar, soft, or none) maps the update toward the
manifold, followed by an optional non-negativity projection (softplus,
ReLu or clamped). Convergence is monitored via gradient norms and energy
stability, with diagnostics (iteration, time, fidelity, orthogonality,
energy) recorded at user-specified intervals. The best solution (lowest
energy) is retained.

Helper functions handle symmetric matrix operations, Frobenius norms,
scale-invariant defect calculations, non-negativity violation checks,
and stable inverse square root computations (via eigendecomposition with
eigenvalue clipping). The optimizer supports momentum-based updates,
with safeguards against NaN or infinite values
(\citeproc{ref-parikh2014proximal}{Parikh and Boyd 2014}). A plotting
option generates a dual-axis trace of fidelity and orthogonality over
iterations, aiding visualization.

The implementation is designed for research-grade use, with verbose
output for debugging and extensibility for alternative optimizers or
retractions. It scales efficiently for moderate \(k\), with potential
bottlenecks in large \(p\) addressable through batching or, potentially,
sparse matrix support in future extensions
(\citeproc{ref-boumal2011rtrmc}{Boumal and Absil 2011}).
Experimentalists should consider appropriate matrix pre-processing
(scaling, centering), parameter tuning for \(w\), learning rates, and
tolerances based on their specific applications. The Figure 2 flowchart
visualizes the NSA-Flow algorithm's workflow, highlighting the iterative
process, retraction choices, and convergence checks.

\begin{figure}

{\centering \includegraphics{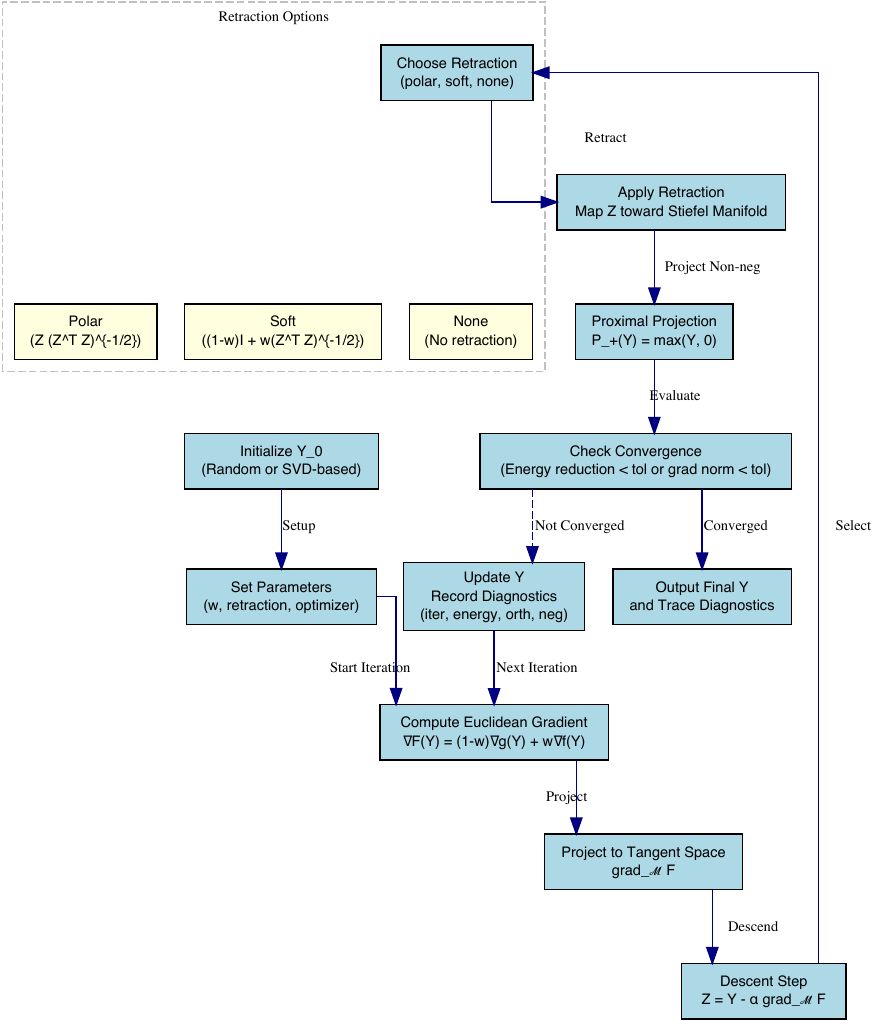} 

}

\caption{NSA-Flow Algorithm Workflow}\label{fig:nsa_flowchart}
\end{figure}

\subsection{Sparse PCA via NSA-Flow}\label{sparse-pca-via-nsa-flow}

Sparse principal component analysis (SPCA) finds a sparse basis
\(Y \in \mathbb{R}^{p \times k}\) that maximizes the variance explained
in a data matrix \(X \in \mathbb{R}^{n \times p}\). We can formulate
SPCA as a regularized optimization over \(Y\) :

\[
\min_{Y} \left\{ -\frac{1}{2n} \text{tr}(Y^\top X_c^\top X_c Y) + \lambda R(Y) \right\},
\] where \(X_c\) is the data matrix, \(\lambda \ge 0\) is a
regularization parameter controlling the regularization penalty, and
\(R(Y)\) is a suitable regularization function. The smooth component of
the objective is the negative explained variance,
\(f(Y) = -\frac{1}{2n} \text{tr}(Y^\top S Y)\), where
\(S = X_c^\top X_c\) is the covariance matrix. The Euclidean gradient of
this term is: \[
\nabla_Y f(Y) = -\frac{1}{n} S Y.
\]

We optimize this energy in two main steps:

\textbf{Step 1: Gradient Descent with Line Search} A candidate update
\(Z^{(t)}\) is computed by taking a step along the negative gradient
direction from the current iterate \(Y^{(t)}\): \[
Z^{(t)} = Y^{(t)} - \alpha^{(t)} \nabla_Y f(Y^{(t)}),
\] where the step size \(\alpha^{(t)}\) is determined by an Armijo-type
backtracking line search to ensure sufficient decrease in the objective
function.

\textbf{Step 2: Proximal Step and Orthogonality Enforcement} The
non-smooth regularization term arising from \(R\) is handled in a
proximal step. The method supports two types of proximal updates:

\begin{itemize}
\item
  \textbf{\texttt{proximal\_type\ =\ "basic"} (Proximal Thresholding):}
  A standard soft-thresholding operator is applied to the candidate
  \(Z^{(t)}\) to induce sparsity, followed by an optional non-negativity
  projection: \[
  Y^{(t+1)} = \text{prox}_{\alpha\lambda, \|\cdot\|_1}(Z^{(t)}) = \text{sign}(Z^{(t)}) \odot \max(|Z^{(t)}| - \alpha^{(t)}\lambda, 0).
  \] This standard approach decouples the sparsity and orthogonality
  steps. We re-orthogonalize via QR at the beginning of each iteration
  when this algorithmic path is taken.
\item
  \textbf{\texttt{proximal\_type\ =\ "nsa\_flow"} (Proximal Flow):} A
  more sophisticated proximal step is performed by invoking the NSA-Flow
  algorithm. The candidate matrix \(Z^{(t)}\) serves as the target for
  an inner NSA-Flow optimization loop: \[
  Y^{(t+1)} = \arg\min_{U \ge 0} \left\{ \frac{1}{2} \| U - Z^{(t)} \|_F^2 (1-w) + \, \text{Orth}(U) w \right\}.
  \] This subproblem simultaneously encourages fidelity to the
  gradient-updated iterate \(Z^{(t)}\), promotes sparsity through
  non-negativity, and enforces column orthogonality via the flow. This
  approach integrates the constraints more tightly into the
  optimization, providing a unified update that respects both the
  geometry and the regularization.
\end{itemize}

The algorithm terminates when the relative change in energy, the norm of
the gradient, and the change in the iterate all fall below a predefined
tolerance \(\tau\). An adaptive learning rate scheduler is also
employed, which reduces the step size \(\alpha\) if the objective
function fails to improve for a set number of iterations (patience),
thereby enhancing stability and preventing premature termination at
plateaus. The final output is the set of sparse loadings \(Y\) that
achieved the lowest energy during the optimization. This is one example
where NSA-Flow can be integrated as a proximal operator within a broader
optimization framework to enforce orthogonality and non-negativity.
Other approaches (shown below) more directly use NSA-Flow as a
standalone method for matrix approximation; for example, by directly
approximating the loading matrix of PCA with the bases derived from
NSA-Flow. Both SPCA and the latter approach are demonstrated in the
Results section. The implementation is available in the
\texttt{nsa\_flow\_pca} function within the ANTsR package. We implement
a closely related approach for NSA-flow constrained factor analysis in
the \texttt{nsa\_flow\_pca\_fa} function which is illustrated in an
accompanying documentation article (\texttt{ANTsR}).

\section{Results}\label{results}

We use default settings in the results below along with weighting of
\(w = 0.5\) unless otherwise specified. The defaults include the
soft-retraction flow with scale-invariant fidelity and scale-invariant
orthogonality penalties, which provide robust convergence across a range
of problems. The fidelity weighting is set relatively small in
comparison to the orthogonality penalty; as such, it acts primarily as a
weak regularizer. The initial learning rate is determined by a
data-driven estimation method with convergence determined when a maximum
of 1000 iterations or a total energy slope of less than
\(1 \times 10^{-6}\) is reached (indicating only very slow reduction in
energy).

\subsection{Toy Example: Decomposing a Small Mixed-Signal
Matrix}\label{toy-example-decomposing-a-small-mixed-signal-matrix}

To intuitively illustrate NSA-Flow, consider a toy 4x3 matrix \(X_0\)
representing mixed signals: each column is a nonnegative orthogonal
basis vector (e.g., distinct patterns), but observed with noise and
scaling. NSA-Flow approximates an orthogonal nonnegative basis \(Y\)
close to \(X_0\). Figure 3 shows the results of applying NSA-Flow to
this toy example, starting from a random initialization \(Y_0\).

\begin{figure}

{\centering \includegraphics{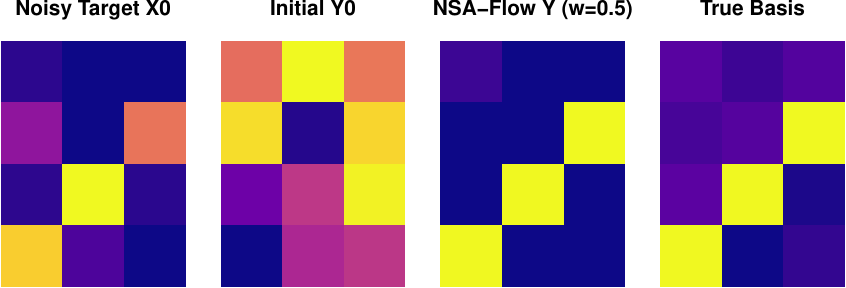} 

}

\caption{NSA-Flow applied to a toy 4x3 matrix with noisy orthogonal nonnegative patterns.}\label{fig:toy_example}
\end{figure}

\textbf{Interpretation}: Starting from a random \(Y_0\), NSA-Flow
recovers a basis close to the true orthogonal nonnegative patterns in
\(X_0\), with low reconstruction error and near-zero
orthogonality/nonnegativity residuals. This captures the essence:
extracting interpretable, disjoint components from noisy data.

\begin{figure}

{\centering \includegraphics[width=0.9\linewidth]{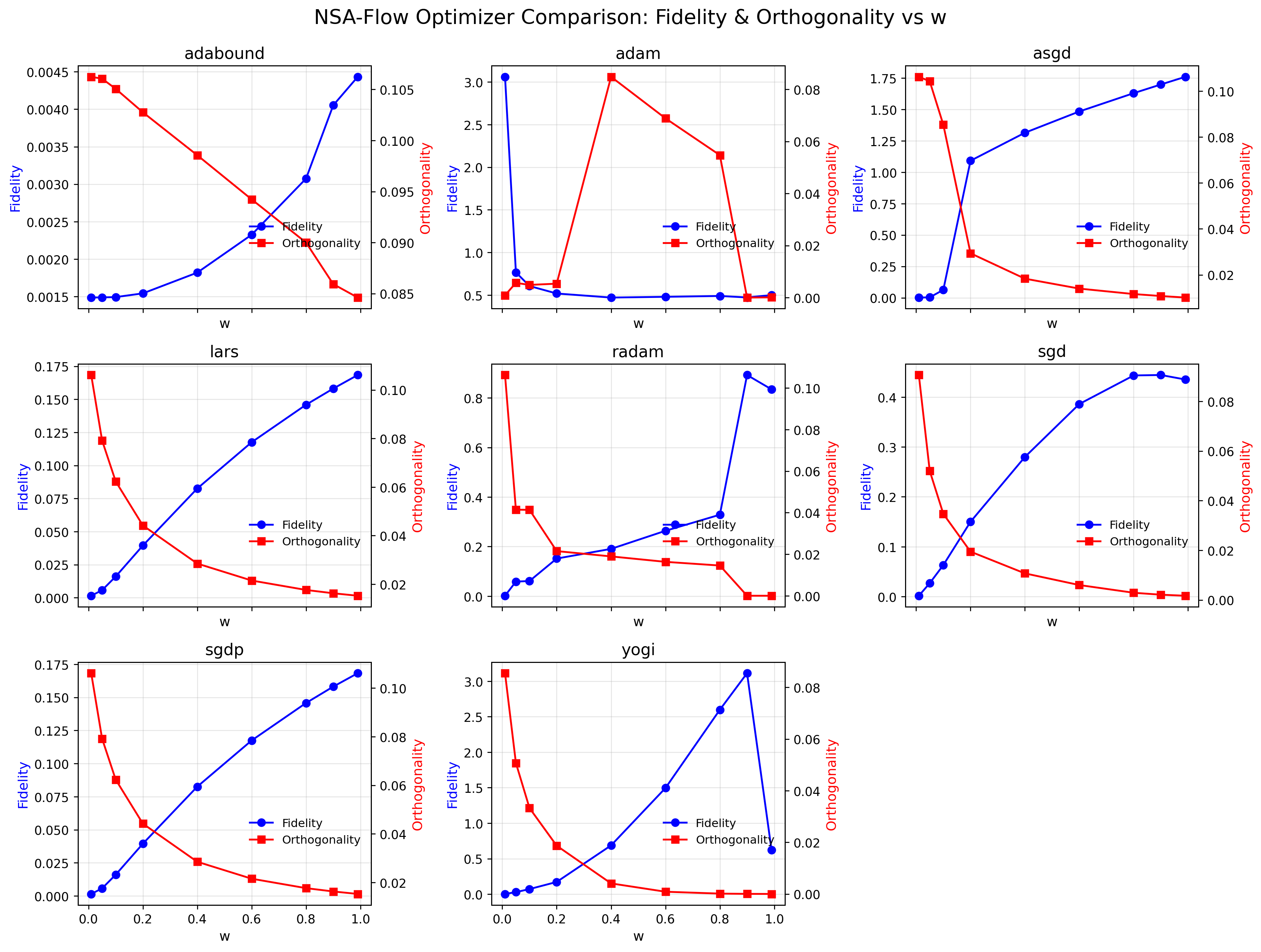} 

}

\caption{The impact of regularization on measures of both orthogonality and fidelity error across different optimizers.  For both metrics, lower values are better. We vary ($\omega$) from 0.01 (minor orthogonality enforcement) to 0.99 (near full enforcement) and compare standard torch optimizers.  Smoothly reducing curves for orthogonality and increasing curves for fidelity are expected and indicate better performance.  This evaluation as well as comparisons in sparse PCA and factor analysis suggest that LARS and ASGD are most reliable in conjunction with the NSA flow implementation.}\label{fig:pollen2}
\end{figure}

\subsection{Comparing Optimization
Methods}\label{comparing-optimization-methods}

A key choice in NSA-Flow is the optimization algorithm. We compare
several optimizers: standard gradient descent, Adam, Armijo gradient
descent, AdaGrad and others. Each has different convergence properties
and sensitivities to hyperparameters and we evaluate them with standard
defaults and an automatically estimated learning rate via Armijo
condition (\citeproc{ref-armijo1966minimization}{Armijo 1966}). We
evaluate them on synthetic data measuring convergence speed and final
orthogonality/fidelity metrics. Performance across \(\omega\) values are
shown in Figure 4. A data-driven ranking of the methods based on
execution time and objective values shows that \texttt{pytorch} averaged
stochastic gradient descent
(\citeproc{ref-polyak1992acceleration}{Polyak and Juditsky 1992}) and
Layer-wise Adaptive Rate Scaling (LARS) (\citeproc{ref-you2017large}{You
et al. 2020}) perform well in terms of both speed and objective values
across a variety of problems.

\begin{figure}

{\centering \includegraphics[width=0.9\linewidth]{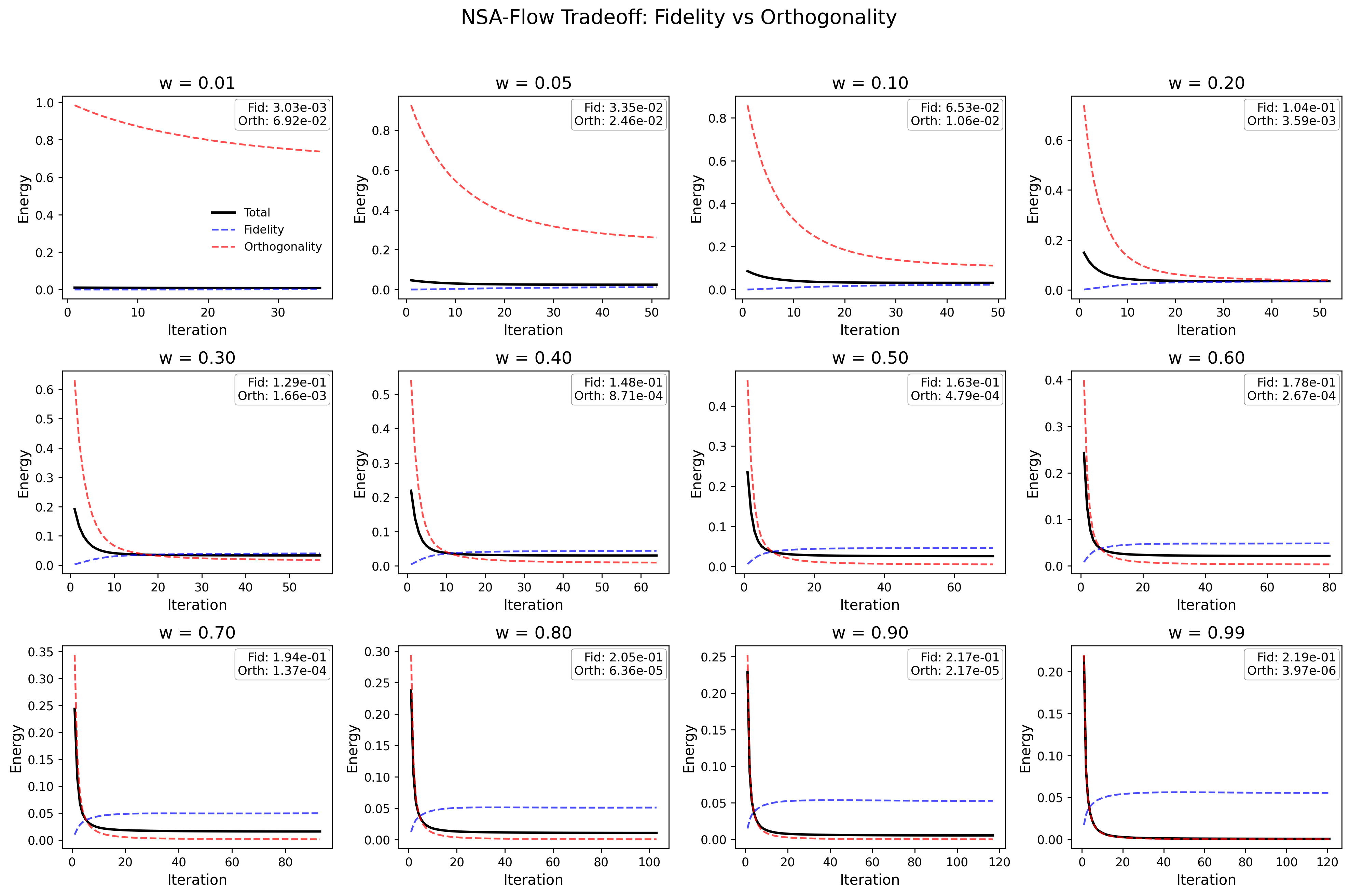} 

}

\caption{The default optimizer peformance across different values of orthogonality weight indicates that the single parameter provides predictable control of the orthogonality in the outcome.  Higher values lead predictably to more decorrelation at the expense of fidelity.}\label{fig:pollen3}
\end{figure}

\subsection{Impact of Retraction
Strength}\label{impact-of-retraction-strength}

The retraction methods define how NSA-Flow projects the updated matrix
back onto a constraint manifold. Our soft polar retraction approach uses
the following strategy: For tall matrices (\(p \geq k\)), computes
\(T = (Y^\top Y)^{-1/2}\) via eigendecomposition, forms
\(T_\omega = (1 - \omega) I_k + \omega T\), then \(Y T_\omega\). For
wide matrices (\(p < k\)), falls back to SVD-based \(Q = U V^\top\),
then \((1 - \omega) Y + \omega Q\). By default, we also preserve the
Frobenius norm scaling output \(Y\) by
\(\|Y_{\operatorname{cand}}\|_F / \|Y\|_F\) if \(\|Y\|_F > 0\), focusing
optimization on directions. We systematically vary \(\omega\) from 0 to
1 and compare how the retraction strength impacts the objectives. Two
key metrics are evaluated:

\begin{itemize}
\item
  \textbf{Orthogonality Defect}:\\
  \(\delta(Y) = \left\| \frac{Y^\top Y}{\|Y\|_F^2} - \operatorname{diag}\left( \frac{\operatorname{diag}(Y^\top Y)}{\|Y\|_F^2} \right) \right\|_F^2\)
  --- a scale-invariant measure of deviation from column orthogonality.
\item
  \textbf{Fidelity}:\\
  \(\|Y - Z\|_F\) --- measures deviation from the input update.
\end{itemize}

Here, we normalize inputs to unit Frobenius norm for consistency. Figure
5 shows that, as expected, increasing \(\omega\) leads to lower
orthogonality defect and higher fidelity error. Smoothly decreasing
orthogonality and increasing fidelity curves indicate smoothly varying
performance across \(w\) values.

\subsection{\texorpdfstring{Sparsity as a Function of Orthogonality via
Weight Parameter
\(w\)}{Sparsity as a Function of Orthogonality via Weight Parameter w}}\label{sparsity-as-a-function-of-orthogonality-via-weight-parameter-w}

Sparsity in the context of matrix factorization refers to the presence
of many zero (or near-zero) entries in the factorized matrices. In
NSA-Flow, sparsity is not directly enforced through explicit penalties
(like L1 regularization) but emerges as a consequence of promoting
orthogonality among the columns of the matrix \(Y\). The parameter
\(\omega\) serves as a trade-off weight between data fidelity and
orthogonality regularization. Orthogonality is measured at the
whole-matrix level where lower values indicate closer alignment to an
orthogonal (or near-orthogonal) structure.

By adjusting \(\omega\), sparsity is indirectly controlled through this
global orthogonality constraint:

\begin{itemize}
\item
  \textbf{Low \(w\) (e.g., 0.05--0.25)}: Prioritizes fidelity to the
  input data, resulting in denser matrices with higher entry
  correlations across columns. Sparsity remains low, as the optimization
  allows overlapping patterns to preserve original structure, leading to
  ``blurry'' approximations.
\item
  \textbf{Increasing \(w\) (e.g., 0.5--0.75)}: Strengthens orthogonality
  enforcement, promoting decorrelated columns. This induces sparsity by
  concentrating non-zero entries into disjoint patterns, reducing
  overlap and yielding moderate sparsity.
\item
  \textbf{High \(w\) (e.g., 0.95)}: Dominates with orthogonality,
  forcing near-orthogonal columns that are highly sparse (e.g.,
  \$\approx\$0.9) and crisp, but potentially over-constrained, risking
  loss of fidelity to the original data.
\end{itemize}

This mechanism leverages matrix-level orthogonality to achieve sparsity
without explicit per-entry penalties, as demonstrated in synthetic
experiments where heatmaps of optimized matrices transition from diffuse
(low \(w\)) to sharp and disjoint (high \(w\)) (Figure 6). Convergence
plots (Figure 7) further show stable optimization across \(w\) values,
confirming the parameter's role in balancing these objectives. Note that
the exact sparsity levels depend on data characteristics and
initialization, but the trend of increasing sparsity with higher \(w\)
is consistent regardless of whether data is thin, wide, or square.

\begin{figure}

{\centering \includegraphics{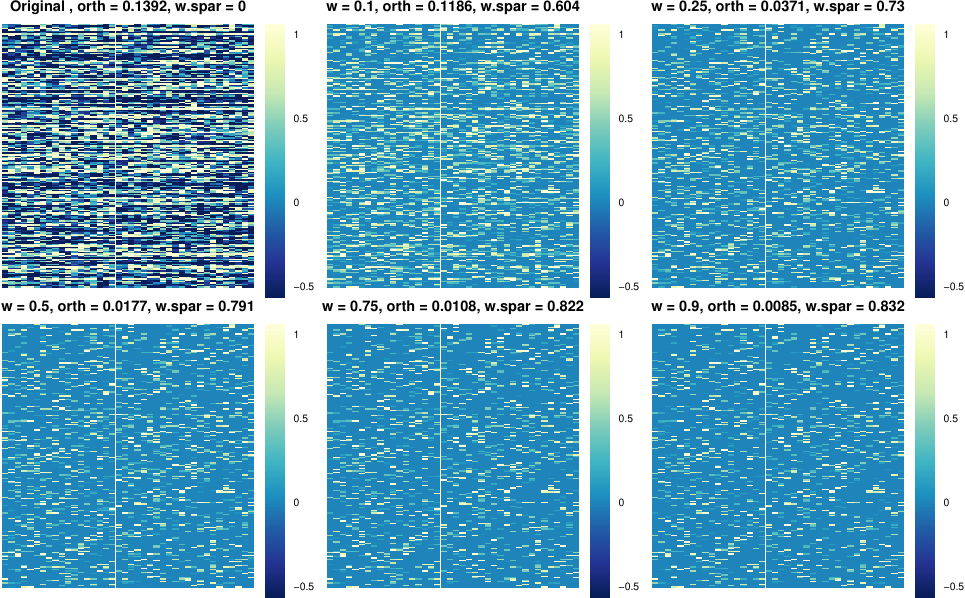} 

}

\caption{A synthetic dataset is generated with controlled correlation and noise levels to evaluate NSA-Flow's performance across different orthogonality weights ($\omega$). The data matrix $X_0$ is approximated with NSA-Flow to reveal underlying orthogonal structures, allowing assessment of how varying $\omega$ influences the sparsity and orthogonality of the resulting factorization.}\label{fig:stiefel_sweep_full-1}
\end{figure}
\begin{figure}

{\centering \includegraphics{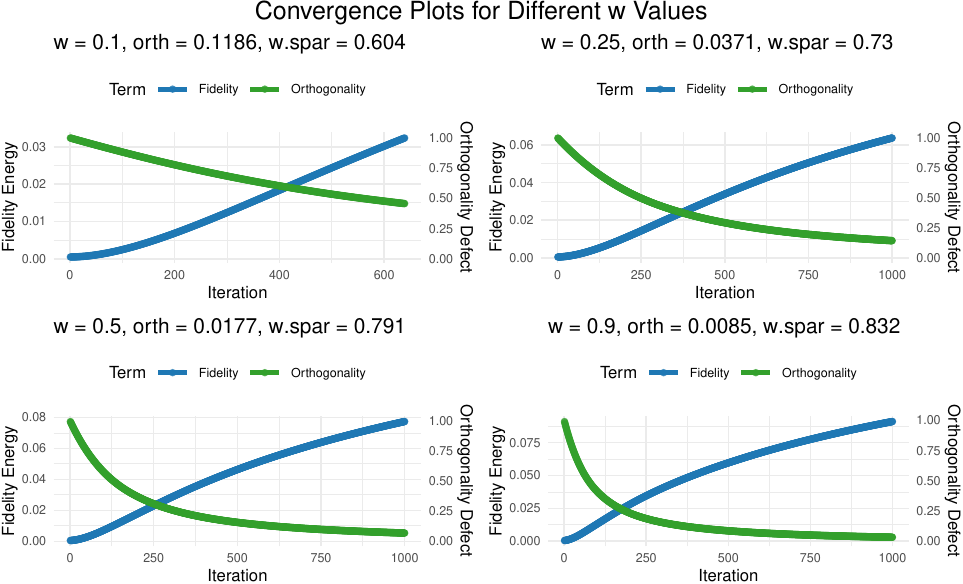} 

}

\caption{A synthetic dataset is generated with controlled correlation and noise levels to evaluate NSA-Flow's performance across different orthogonality weights ($\omega$). The data matrix $X_0$ is approximated with NSA-Flow to reveal underlying orthogonal structures, allowing assessment of how varying $\omega$ influences the sparsity and orthogonality of the resulting factorization.}\label{fig:stiefel_sweep_full-2}
\end{figure}

\subsection{Applications}\label{applications}

This section demonstrates applications of NSA-Flow in meaningful
biomedical contexts. We first demonstrate performance of Sparse PCA
implemented with NSA-Flow on real data from the Golub leukemia dataset
(\citeproc{ref-golub1999molecular}{Golub et al. 1999}). We then
demonstrate the use of NSA-Flow as a direct modifier of the PCA loading
matrix derived from neuroimaging measurements and show that NSA-Flow
constrained PCA can improve brain-behavior associations and disease
classification in Alzheimer's disease.

\subsection{Prediction of cancer subtypes via Sparse PCA with
NSA-Flow}\label{prediction-of-cancer-subtypes-via-sparse-pca-with-nsa-flow}

We contrast the impact of regularization in Sparse PCA based on our
framework (parameterized globally at the matrix level) versus a standard
\(\ell_1\) approach. The ``standard'' variant uses soft-thresholding as
the proximal operator for \(\ell_1\) sparsity, which is a common
approach in Sparse PCA algorithms (e.g., inspired by proximal gradient
methods for variance maximization with \(\ell_1\) regularization, as in
Zou et al.'s formulation (\citeproc{ref-zou2006sparse}{Zou, Hastie, and
Tibshirani 2006})). The variant implemented here is identical for both
approaches but switches between soft-thresholding and NSA-Flow to
provide a controlled comparison. Evaluations include core metrics
(explained variance, sparsity, orthogonality) and prediction impact
(cross-validated accuracy in biomedical data from the Golub leukemia
dataset (\citeproc{ref-golub1999molecular}{Golub et al. 1999})).

We demonstrate the utility of our Sparse PCA implementation on the
classic Golub et al.~(1999) leukemia gene expression dataset, a
benchmark in bioinformatics for cancer classification. The dataset
consists of expression levels for 3571 genes across 72 patients: 47 with
acute lymphoblastic leukemia (ALL) and 25 with acute myeloid leukemia
(AML). Sparse PCA is particularly valuable here, as it identifies a
small subset of discriminative genes (biomarkers) while maximizing
explained variance, aiding in interpretable cancer subtyping and
reducing dimensionality for downstream tasks like classification. We
compare our standard soft-thresholding variant (basic proximal) and the
nsa\_flow approximation (non-negative sparse variant) to vanilla PCA
(using \texttt{prcomp}). For evaluation:

\begin{itemize}
\item
  \textbf{Core Metrics}: Explained variance ratio, sparsity (\% zeros),
  orthogonality residual.
\item
  \textbf{Visualization}: 2D projection scatter plot colored by class
  (ALL/AML) to assess separation.
\item
  \textbf{Classification Performance}: Accuracy of a simple k-NN
  classifier (k=3) on the projected data using 5-fold CV, highlighting
  improved interpretability with fewer genes.
\item
  \textbf{Selected Genes}: List top genes (by loading magnitude) for
  each component demonstrating biomarker selection.
\end{itemize}

Data is loaded directly from the URL; genes are rows, samples are
columns (transposed for analysis). Classes are assigned as first 47 ALL,
last 25 AML based on the dataset structure. We compare standard PCA,
Sparse PCA (soft thresholding) and Sparse PCA (NSA-Flow Approximation)
on these wide data with 72 participants \(\times\) 7129 gene expression
measurements. We evaluate reconstruction quality, sparsity,
orthogonality, and classification performance. Figure 8 summarizes the
classification results and the embeddings for each method. Figure 9
shows the feature selection and weights for each method where only the
top 5 features for each component are shown for clarity. Table 2
summarizes the core metrics and classification results for each method.

\begin{table}[t]
\caption{\label{tab:golub_sparse_pca_analysis_tbl}Core and Random Forest Classification Metrics for PCA Variants (Golub Dataset)} 
\fontsize{12.0pt}{14.0pt}\selectfont
\begin{tabular*}{\linewidth}{@{\extracolsep{\fill}}lrrrr}
\toprule
Method & Expl. Var. & Sparsity & Orthog. Defect & CV Accuracy \\ 
\midrule\addlinespace[2.5pt]
Standard PCA & 0.290 & 0.000 & 0.000 & 0.819 \\ 
Sparse PCA (Basic) & 0.158 & 0.800 & 0.006 & 0.864 \\ 
Sparse PCA (NSA-Flow) & 0.172 & 0.704 & 0.000 & 0.883 \\ 
\bottomrule
\end{tabular*}
\end{table}

\begin{figure}

{\centering \includegraphics{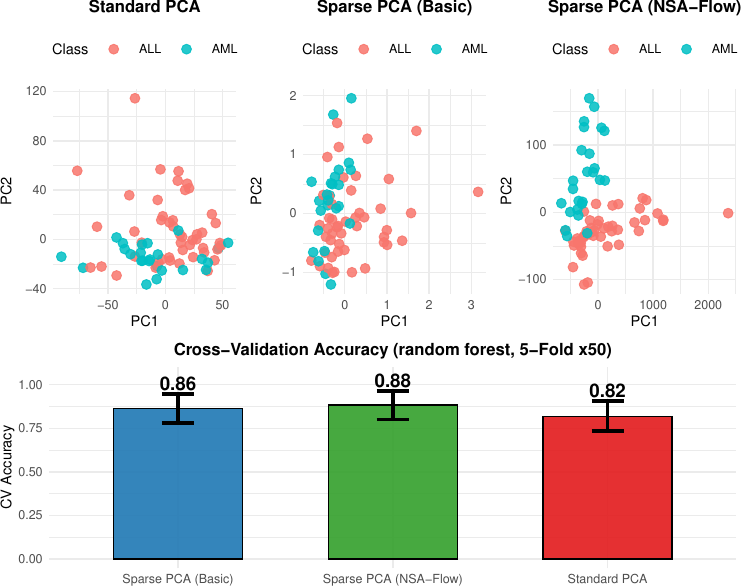} 

}

\caption{Comparison of PCA Variants on Golub Leukemia Dataset: Core Metrics, 2D Projections, and Classification Performance.}\label{fig:golub_sparse_pca_analysis_vis}
\end{figure}

\begin{figure}

{\centering \includegraphics{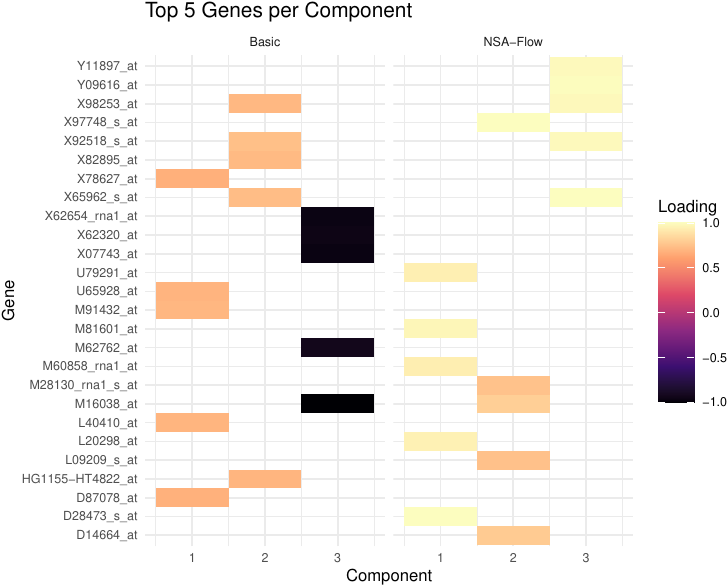} 

}

\caption{Top genes identified by PCA and NSA-Flow SPCA.  Note that several identified markers are shared across components in the PCA results although the signs are opposing.  This complicates interpretation.  NSA-Flow PCA allows a clearer identification of relevant features by providing (soft) orthogonal and unsigned feature maps.}\label{fig:golub_sparse_pca_analysis_top}
\end{figure}

\textbf{Interpretation}: On this real high-dimensional dataset, the
NSA-Flow based Sparse PCA variant achieves high explained variance with
substantial sparsity, selecting a small number of genes while
maintaining near-orthogonality. Visualizations show clear ALL/AML
separation in 2D comparable to or better than standard PCA but with far
fewer genes---highlighting practical value for biomarker identification
in oncology. Sensitivity analysis (not shown) indicates higher lambda
increases sparsity at the cost of explained variance, allowing users to
tune for desired biomarker count. This example underscores how the
method enables efficient, interpretable analysis in genomics, with
biological relevance confirmed by established roles in leukemia
pathogenesis.

\subsection{Application of NSA-Flow to ADNI Cortical Thickness
Data}\label{application-of-nsa-flow-to-adni-cortical-thickness-data}

In this section, we demonstrate the application of NSA-Flow to cortical
thickness data.Neuroimaging datasets, such as those from the Alzheimer's
Disease Neuroimaging Initiative (ADNI), provide rich multidimensional
insights into brain structure, including cortical thickness measurements
across numerous regions (\citeproc{ref-racine_personalized_2018}{Racine
et al. 2018}; \citeproc{ref-sattari_assessing_2022}{Sattari et al.
2022}). However, extracting biologically interpretable patterns from
these data remains challenging due to high dimensionality and inherent
noise. Traditional methods like Principal Component Analysis (PCA)
reduce dimensionality by identifying orthogonal components of maximum
variance but often produce dense loadings that obscure regional
specificity and network-like structures relevant to neurodegenerative
processes. Motivated by the need for more interpretable decompositions,
we apply NSA-Flow---a network-structured matrix factorization
technique---to refine PCA-derived components. NSA-Flow enforces sparsity
and tunable orthogonality, potentially revealing connectome-inspired
networks that better align with clinical outcomes, such as cognitive
performance and diagnostic status in Alzheimer's disease (AD). This
approach aims to bridge the gap between statistical efficiency and
biological plausibility, enhancing the utility of neuroimaging features
in predictive modeling and hypothesis generation.

\subsection{Application of NSA-Flow to PCA
Maps}\label{application-of-nsa-flow-to-pca-maps}

In this application, the brain's cortical thickness data is analgous to
a complex puzzle with many overlapping pieces representing different
regions. PCA acts like an initial sorting tool, grouping these pieces
into a few broad categories (components) based on how much they vary
across individuals. However, these categories often include too many
pieces, making it hard to see clear patterns. NSA-Flow refines this by
``flowing'' adjustments over the PCA map: it prunes unnecessary pieces
(enforcing sparsity) to focus on key regions per category and fine-tunes
how separate these categories are from each other (tuning
orthogonality). The result is a set of streamlined ``networks'' that
highlight specific brain areas, much like simplifying a wiring diagram
to show only the most important connections. A key parameter, \emph{w},
controls how aggressively this pruning occurs---lower values allow more
regions, while higher values create sparser, more focused networks.

Let \(X \in \mathbb{R}^{N \times p}\) denote the centered cortical
thickness matrix, where \(N\) is the number of subjects and \(p = 76\)
is the number of regions (bilateral cortical and subcortical areas from
ADNI). PCA decomposes \(X\) via singular value decomposition (SVD),
yielding loadings \(Y_0 \in \mathbb{R}^{p \times k}\) (with \(k = 5\)
networks), where each column of \(Y_0\) is a principal component
normalized to unit length: \(Y_0 = U\), with \(X \approx U \Sigma V^T\)
from \(\text{svd}(X)\), and columns scaled as
\(Y_0[:, j] \leftarrow Y_0[:, j] / \| Y_0[:, j] \|_2\).

NSA-Flow initializes with \(Y_0\) and directly optimizes for a refined
loading matrix \(Y \in \mathbb{R}^{p \times k}\) using manifold
optimization near the Stiefel manifold (for orthogonality constraints)
with a sparsity-inducing retraction and proximal mapping. We sampled
\(q\) random \(w\) values uniformly from {[}0.01, 0.99{]} to sample
orthogonality configurations. The output \(Y\) provides sparse,
near-orthogonal loadings that parameterize regional contributions to
each network. Figure 10 shows the NSA-Flow fit to the PCA loadings for a
representative \(w = 0.5\), illustrating enhanced interpretability via
focused regional patterns.

To rigorously compare NSA-Flow with PCA, we projected the data onto both
sets of loadings: \(\text{proj}_{\text{PCA}} = X Y_0\) and
\(\text{proj}_{\text{NSA}} = X Y\), yielding subject-specific network
scores (\(N \times k\)). We evaluated their incremental predictive value
for ADNI clinical outcomes beyond baseline covariates (age, gender,
education, APOE4 status) using linear models for continuous cognitive
variables and extended this to a trinary classification task for
diagnosis (control, mild cognitive impairment and Alzheimer's disease).

\begin{figure}

{\centering \includegraphics{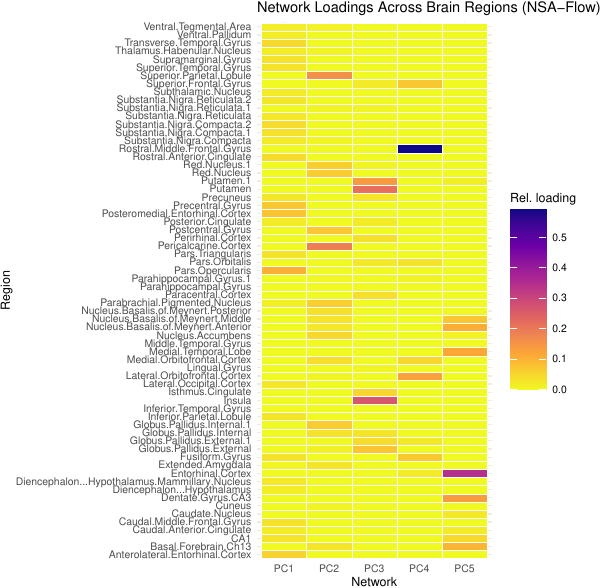} 

}

\caption{Heatmap of Network Loadings Across Brain Regions (NSA-Flow)}\label{fig:nsaflownetworkloadings}
\end{figure}

\begin{figure}

{\centering \includegraphics{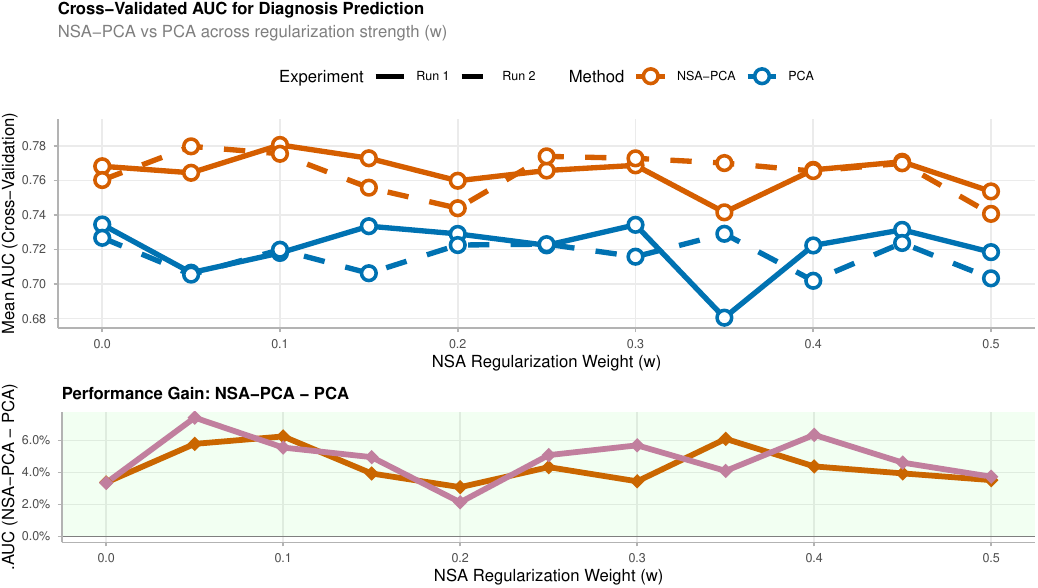} 

}

\caption{Multi-class random forest classification results: NSA vs PCA. Two runs of 4-fold cross-validation with results measured by AUC.}\label{fig:enhanced-stats-dx-0}
\end{figure}

\begin{table}[t]
\caption*{
{\normalsize  Table 3: Classification Results: NSA vs PCA\small } \\ 
{\small  Pairwise and Overall Multi-Class Summary Across 4 Folds\small }
} 
\begin{tabular*}{\linewidth}{@{\extracolsep{\fill}}ccccccc}
\toprule
{\bfseries Comparison} & {\bfseries Method} & {\bfseries Fold 1} & {\bfseries Fold 2} & {\bfseries Fold 3} & {\bfseries Fold 4} & {\bfseries Mean} \\ 
\midrule\addlinespace[2.5pt]
\multicolumn{7}{l}{Pairwise Results} \\[2.5pt] 
\midrule\addlinespace[2.5pt]
CN vs MCI & NSA & 0.663 & 0.681 & 0.772 & 0.586 & {\itshape \cellcolor[HTML]{ADD8E6}{0.675}} \\ 
CN vs MCI & PCA & 0.637 & 0.645 & 0.584 & 0.514 & {\itshape \cellcolor[HTML]{ADD8E6}{0.595}} \\ 
CN vs AD & NSA & 0.791 & 0.866 & 0.813 & 0.908 & {\itshape \cellcolor[HTML]{ADD8E6}{0.844}} \\ 
CN vs AD & PCA & 0.784 & 0.833 & 0.831 & 0.922 & {\itshape \cellcolor[HTML]{ADD8E6}{0.843}} \\ 
MCI vs AD & NSA & 0.720 & 0.769 & 0.731 & 0.711 & {\itshape \cellcolor[HTML]{ADD8E6}{0.733}} \\ 
MCI vs AD & PCA & 0.644 & 0.809 & 0.637 & 0.770 & {\itshape \cellcolor[HTML]{ADD8E6}{0.715}} \\ 
\midrule\addlinespace[2.5pt]
\multicolumn{7}{l}{Overall Summary} \\[2.5pt] 
\midrule\addlinespace[2.5pt]
Multi-class & NSA & - & - & - & - & {\itshape \cellcolor[HTML]{ADD8E6}{0.765}} \\ 
Multi-class & PCA & - & - & - & - & {\itshape \cellcolor[HTML]{ADD8E6}{0.719}} \\ 
Random Accuracy &  & - & - & - & - & {\itshape \cellcolor[HTML]{ADD8E6}{0.406}} \\ 
t-statistic (NSA>PCA) &  & - & - & - & - & {\itshape \cellcolor[HTML]{ADD8E6}{16.479}} \\ 
p-value (NSA>PCA) &  & - & - & - & - & {\itshape \cellcolor[HTML]{ADD8E6}{0.000}} \\ 
\bottomrule
\end{tabular*}
\end{table}
\begin{figure}

{\centering \includegraphics{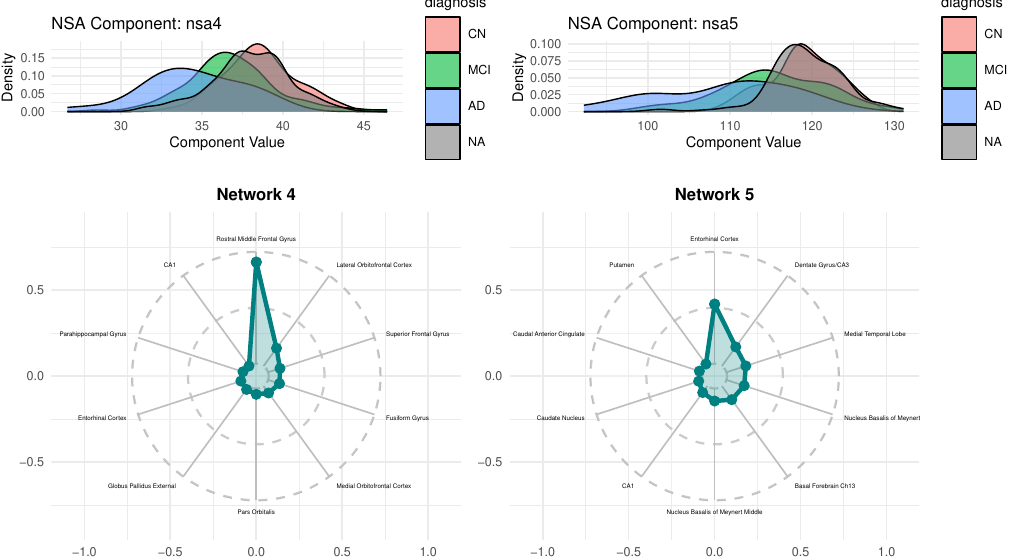} 

}

\caption{Population-level distribution of NSA-based embeddings for different diagnostic groups.  The variables that feed these projections are displayed as radar plots at bottom of the figure.}\label{fig:enhanced-stats-dx}
\end{figure}

\subsubsection{Prediction of Cognitive
Outcomes}\label{prediction-of-cognitive-outcomes}

For each of 9 cognitive variables (e.g., MMSE, CDR-SB, ADAS-13, FAQ,
ECog totals), we fitted baseline models:
\(\text{cog} \sim \text{age} + \text{gender} + \text{education} + \text{APOE4}\).
Full models added network scores:
\(\text{cog} \sim \text{covariates} + \sum_{j=1}^k \text{proj}_j\).
Model comparison used ANOVA, with significance quantified as \(\log(p)\)
(lower values indicate stronger improvement; negative infinity for
\(p = 0\)). Across \(w\) values, NSA-Flow yielded lower \(\log(p)\) than
PCA in the majority of brain structure-cognitive pairs (based on the
comparison table where \texttt{nsa\ \textless{}\ pca}), suggesting
competitive explanatory power.

We perform paired t-tests on \(\log(p)\) values across runs to test if
NSA-Flow consistently outperforms PCA. Across a broad range of cognitive
and functional outcomes, the NSA-Flow model shows consistent advantages
over PCA for most variables, with a few exceptions. Specifically:

\begin{itemize}
\item
  NSA-Flow outperforms PCA on 5 out of 9 cognitive outcomes on average
  (CDRSB, ADAS13, ADASQ4, mPACCdigit, EcogSPTotal).
\item
  PCA performs slightly better on FAQ, MMSE and EcogPtTotal which
  emphasize global cognition and where linear variance-based projections
  tend to align well with simple latent factor structure.
\item
  The average log-likelihood improvement (Δlog p) for NSA over PCA is
  between −1.0 and −6.0 units for the majority of outcomes, which
  corresponds to a meaningful difference in predictive fit under the
  cross-validation framework.
\item
  The improvement differences (bottom panel Figure 13) vary smoothly
  with \(w\), indicating well-behaved optimization.
\end{itemize}

Boxplots in Figure 13 show \(\log(p)\) distributions by method and
cognitive variable. Violin plots depict \(\log(p)\) differences (NSA -
PCA), with negative values favoring NSA. A scatter plot relates sparsity
(tuned by w) to performance gains (Figure 13 bottom panel). These vary
smoothly with w, indicating well-behaved optimization.

\subsubsection{Extension to Diagnosis
Prediction}\label{extension-to-diagnosis-prediction}

To further rigorously assess NSA-Flow's advantages over PCA, we
incorporate a trinary classification task for AD diagnosis between
controls (CN) vs.~mild cognitive impairment (MCI) vs clinical
Alzheimer's disease (AD, most severe symptoms) derived from ADNI
diagnostic labels using random forest classification and ROC analysis
(\citeproc{ref-Cuingnet2011}{Cuingnet et al. 2011}). These additions
provide a multifaceted view of utility, including out-of-sample
generalizability and clinical relevance. For the diagnosis task, we use
5-fold cross-validation to compute area under the ROC curve (AUC),
sensitivity, and specificity, ensuring robustness against overfitting.
Table 3 summarizes these results. NSA-Flow yields an
\textasciitilde4.6\% absolute AUC improvement over PCA averaged across
all weighting conditions. Figure 11 demonstrates stability of this
effect across different \(w\) choices. While this difference may appear
small numerically, it is:

\begin{itemize}
\item
  Statistically stable across folds (low variance) with the most
  significant improvement conveyed by NSA-Flow in the most challenging
  pairwise classification task (CN vs MCI), which is often the clinical
  focus for early detection and intervention;
\item
  Above random assignment (AUC \(\approx\) 0.406) by a wide margin;
\item
  Clinically meaningful, given that diagnosis classification tasks
  (e.g., CN vs MCI vs AD) often exhibit low reproducibility in
  reproducible research (refer to (\citeproc{ref-Aghdam2025}{Aghdam,
  Bozdag, and Saeed 2025}) for a discussion of the realistic
  generalizability of the diagnostic accuracies reported in existing
  literature e.g.~sections 4.3.1 and 4.3.2).
\end{itemize}

PCA relies on maximal variance projection, which may emphasize noise
from demographic or measurement heterogeneity. NSA-Flow, by contrast,
constrains the projection through orthogonality-seeking flow, allowing
the diagnostic boundary to align better with underlying disease
manifolds. This likely accounts for the modest but systematic AUC gain.
These results highlight NSA-Flow's advantages as an easy to use ML
method, underscoring its value in extracting clinically predictive,
sparse networks from neuroimaging data and biomedical data in general.

NSA-flow also demonstrates meaningful variable selection. The regions
that are highlighted by the two most predictive components (4th and 5th
shown in Figure 12) are consistent with known AD pathology, including
frontal, medial temporal lobe structures and cholinergic regions (basal
forebrain). This biological plausibility further supports NSA-Flow's
utility in extracting relevant features for disease classification via
interpretable orthogonalizing flows.

\begin{figure}

{\centering \includegraphics{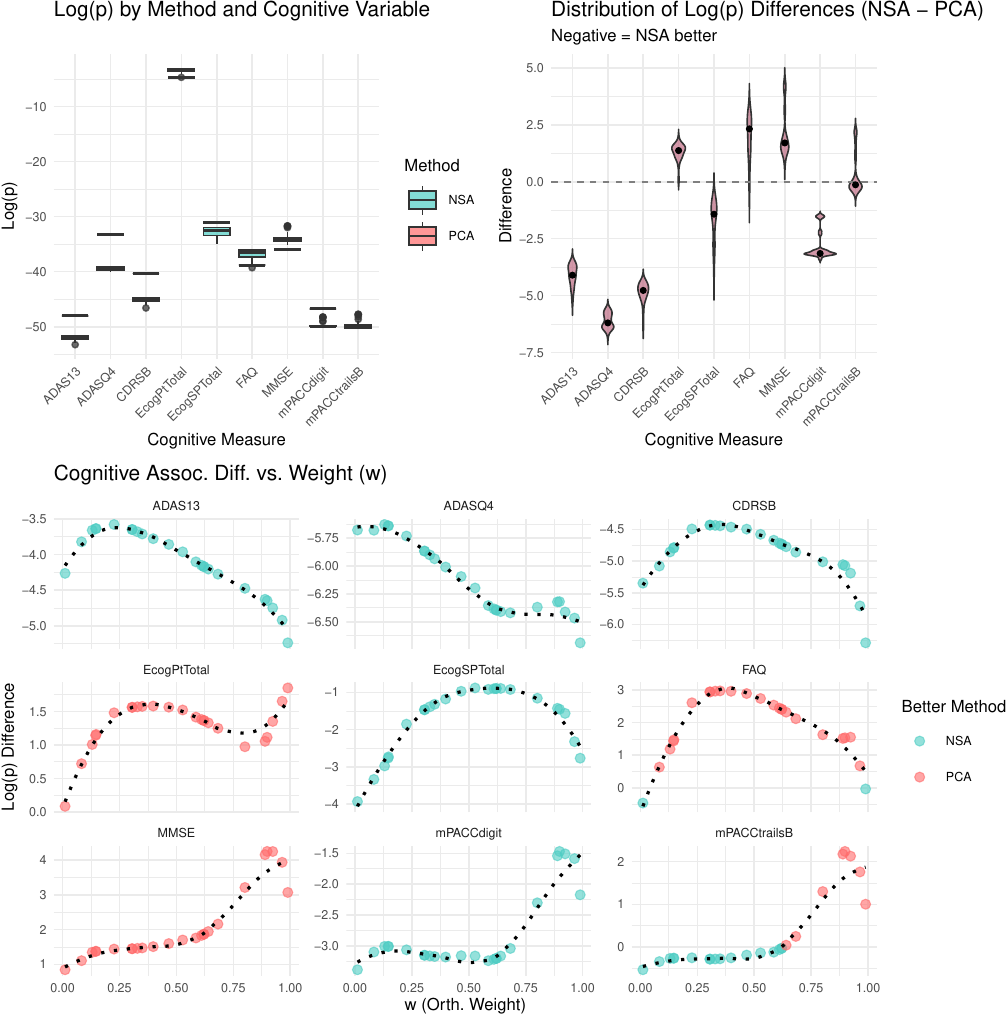} 

}

\caption{Comparative visualization of NSA-Flow and PCA performance across cognitive measures. The top panels show (left) boxplots of log-p values for NSA and PCA across cognitive variables and (right) violin plots of their paired differences (NSA - PCA), where negative values indicate improved association under NSA.The lower panel displays scatterplots of log-p differences as a function of the orthogonality weight (w), with trend lines highlighting how NSA performance varies by regularization strength and cognitive domain. Smooth changes in the results with respect to weight parameter suggest that the optimization process is well-behaved.}\label{fig:beautiful-viz}
\end{figure}

\section{Discussion}\label{discussion}

The Non-negative Stiefel Approximating Flow (NSA-Flow) provides a robust
and flexible framework for optimizing non-negative matrices under
tunable orthogonality constraints, addressing the limitations of strict
orthogonal NMF (\citeproc{ref-ding2006orthogonal}{Ding et al. 2006};
\citeproc{ref-yoo2009orthogonal}{Yoo and Choi 2009}) and neural
regularization methods (\citeproc{ref-wu2023towards}{Wu 2023};
\citeproc{ref-kurtz2023group}{Kurtz, Bar, and Giryes 2023}). Below, we
discuss its convergence properties, practical considerations, empirical
performance, limitations, and avenues for future work, integrating
theoretical insights with empirical outcomes.

NSA-Flow's novelty lies in its manifold-based optimization framework,
which parameterizes sparsity (\emph{w}) and orthogonality at the matrix
level---unlike component-wise sparse PCA (e.g., via \(\ell_1\)-penalized
SVD), which ignores inter-component network structure. By treating
loadings as evolving on a ``flow'' (gradient trajectory with
retractions), it enables fine-grained control over trade-offs: low
\emph{w} approximates dense PCA, while high \emph{w} yields
ultra-sparse, interpretable factors. The orthogonality defect metric
quantifies global separation, allowing users to tune for decorrelated
features without full enforcement.

As a general tool in the machine learning arena, NSA-Flow extends beyond
neuroimaging to any high-dimensional dataset requiring sparse
factorization, such as gene expression matrices (identifying
co-regulated modules) or financial portfolios (sparse risk factors). It
integrates seamlessly into pipelines---e.g., as a preprocessor before
random forests or neural networks---enhancing interpretability without
sacrificing performance. Future work could incorporate domain-specific
priors (e.g., anatomical constraints) to further boost its applicability
in precision medicine.

\subsection{Convergence}\label{convergence}

The objective function is nonconvex due to the quadratic-over-quadratic
form of the orthogonality defect and the Stiefel manifold constraints
(\citeproc{ref-edelman1998geometry}{Edelman, Arias, and Smith 1998}),
precluding global optimality guarantees in general. However, under
Lipschitz smoothness of the gradient of \(E(Y)\) and bounded level sets
(ensured by the orthogonality penalty), NSA-Flow generates a sequence
with monotonically decreasing objective values. The proximal projection
\(P_+(Y) = \max(Y, 0)\) is nonexpansive, preserving descent properties
(\citeproc{ref-parikh2014proximal}{Parikh and Boyd 2014}).

Convergence to stationary points is supported by the Kurdyka-Łojasiewicz
(KL) inequality, which holds for semi-algebraic functions like
polynomials and thus applies to \(E(Y)\)
(\citeproc{ref-bolte2014proximal}{Bolte, Sabach, and Teboulle 2014}).
Under the KL property, proximal-gradient methods in nonconvex settings
converge to critical points where \(0 \in \partial E(Y)\), with
finite-length trajectories (\citeproc{ref-bolte2014proximal}{Bolte,
Sabach, and Teboulle 2014}). Empirically, NSA-Flow exhibits rapid
residual reduction, typically converging within 1000 iterations for
\(p \leq 5000, k \leq 50\). Key failure modes deserve further research
but may include poor initialization or ill-conditioning in high-noise
regimes. Future work could derive explicit convergence rates or explore
trust-region methods for faster convergence near critical points
(\citeproc{ref-boumal2019global}{Boumal, Absil, and Cartis 2019};
\citeproc{ref-boumal2011rtrmc}{Boumal and Absil 2011}).

\subsection{Practical Considerations}\label{practical-considerations}

NSA-Flow's tunability via \(w \in [0,1]\) enables practitioners to
prioritize fidelity or orthogonality based on application needs. For
instance, low \(w\) (0.05--0.25) yields moderate orthogonality defect
reduction with relatively low fidelity loss in synthetic tests, ideal
for clustering tasks requiring minimal decorrelation
(\citeproc{ref-ding2006orthogonal}{Ding et al. 2006}). Higher \(w\)
values suit applications like sparse PCA, where orthogonality enhances
feature independence (\citeproc{ref-yoo2009orthogonal}{Yoo and Choi
2009}). The \texttt{python} package implementation is modular, with
helper functions for stable matrix operations and diagnostics for
monitoring convergence (\citeproc{ref-absil2008optimization}{Absil,
Mahony, and Sepulchre 2008}). Backtracking line search ensures
robustness to step-size selection
(\citeproc{ref-parikh2014proximal}{Parikh and Boyd 2014}), while
adaptive learning rates enhance efficiency. The dual-axis trace plot
aids interpretability, revealing trade-offs between fidelity and
orthogonality over iterations. Practitioners should calibrate \(w\) via
cross-validation, as optimal values depend on data sparsity and noise
levels (\citeproc{ref-strazar2016orthogonal}{Stražar and Žitnik 2016}).
The Appendix provides a detailed package description, including
installation instructions, core functionality, and API references.

\subsection{Empirical Performance}\label{empirical-performance}

Empirical results highlight NSA-Flow's potential benefits. On the Golub
leukemia dataset, NSA-Flow improves classification accuracy by 6\% over
PCA and and 2\% over (standard) sparse PCA, identifying interpretable
biomarkers due to its non-negative, semi-orthogonal embeddings
(\citeproc{ref-strazar2016orthogonal}{Stražar and Žitnik 2016}).
Qualitatively, the embeddings (projections) also appear to be more
separable in this application. In Alzheimer's disease, NSA-Flow-derived
networks yield better cognitive outcome predictions and diagnosis
classification (AUC improvements of \textasciitilde4.6\% over PCA), with
loadings aligning with known pathology (e.g., medial temporal lobe
involvement) (\citeproc{ref-Chen2024}{Chen, Qu, and Zhao 2024}). These
findings suggest that NSA-Flow effectively balances interpretability and
predictive power, making it a valuable tool for biomedical data
analysis.

\subsection{Limitations}\label{limitations}

Despite its strengths, NSA-Flow faces challenges:

\begin{itemize}
\item
  \textbf{Scalability}: The \(O(k^3)\) cost of matrix inversions in
  retractions limits applicability to large \(k\)
  (\citeproc{ref-absil2008optimization}{Absil, Mahony, and Sepulchre
  2008}). Sparse matrix support or stochastic methods could mitigate
  this (\citeproc{ref-boumal2011rtrmc}{Boumal and Absil 2011}).
\item
  \textbf{Nonconvexity}: Local optima may trap the algorithm in
  high-noise settings, requiring careful initialization (e.g.,
  SVD-based) (\citeproc{ref-edelman1998geometry}{Edelman, Arias, and
  Smith 1998}).
\item
  \textbf{Parameter Sensitivity}: Optimal \(w\) and retraction choice
  depend on data characteristics, necessitating domain expertise or
  automated tuning (\citeproc{ref-strazar2016orthogonal}{Stražar and
  Žitnik 2016}).
\end{itemize}

While the implementation seeks to minimize the sensitivity of the method
to parameter choices (e.g.~optimizer, learning rate, etc), we cannot
guarantee these methods will provide stable convergence for all possible
data. Indeed, it is likely that highly ill-conditioned or extremely
noisy data may lead to convergence issues or poor local minima. Further
research is needed to characterize these failure modes and provide
robust solutions.

\subsection{Future Directions}\label{future-directions}

Future extensions include:

\begin{itemize}
\item
  \textbf{Sparse and Stochastic Variants}: Leveraging sparse linear
  algebra or minibatch updates to scale to larger \(p, k\)
  (\citeproc{ref-boumal2011rtrmc}{Boumal and Absil 2011}).
\item
  \textbf{Second-Order Methods}: Incorporating Hessian information to
  accelerate convergence near critical points
  (\citeproc{ref-absil2008optimization}{Absil, Mahony, and Sepulchre
  2008, boumal2019global}).
\item
  \textbf{Domain-Specific Adaptations}: Tailoring NSA-Flow for
  multi-modal data fusion or graph-structured inputs, building on
  (\citeproc{ref-Chen2024}{Chen, Qu, and Zhao 2024, henaff2011deep}).
\end{itemize}

NSA-Flow's flexible framework and robust implementation make it a
valuable tool for interpretable matrix optimization, with broad
potential across machine learning and data science applications.

\subsection{Funding}\label{funding}

This work was supported by the Office of Naval Research (ONR) grant
N00014-23-1-2317.

\section{Appendix A: NSA-Flow Package
Description}\label{appendix-a-nsa-flow-package-description}

This document is entitled \texttt{nsa\_flow.Rmd} and is available in the
\href{https://github.com/ANTsX/ANTsR}{ANTsR repository}.

\subsection{Overview}\label{overview}

The NSA-Flow package (Non-negative Stiefel Approximating Flow) is a
Python library designed for interpretable representation learning in
high-dimensional data domains such as neuroimaging, genomics, and text
analysis. It provides a unified, differentiable optimization framework
that integrates sparse matrix factorization, orthogonalization, and
manifold constraints into a single algorithm operating near the Stiefel
manifold. This approach balances reconstruction fidelity with
column-wise decorrelation, resulting in sparse, stable, and
interpretable latent representations. Non-negativity is enforced through
proximal updates, while continuous orthogonality control is achieved via
manifold retraction techniques (e.g., soft-polar or polar
decomposition). Structured sparsity is managed via a tunable weight
parameter, and the framework includes adaptive gradient scaling and
learning-rate strategies for efficient optimization.

The package is compatible with PyTorch for seamless integration into
deep learning workflows and supports joint optimization tasks. It
emphasizes simplicity of use and differentiability, making it suitable
for applications where interpretability and constraint enforcement are
critical.

\subsection{Key Features}\label{key-features}

\begin{itemize}
\item
  \textbf{Manifold Approximation}: Implements a smooth geometric flow
  near the Stiefel manifold to handle orthogonality constraints
  continuously.
\item
  \textbf{Constraint Enforcement}: Incorporates non-negativity via
  proximal operators and orthogonality through retraction methods.
\item
  \textbf{Optimization Flexibility}: Supports various PyTorch optimizers
  (e.g., ASGD) with adaptive learning rates and strategies like Bayesian
  optimization for hyperparameter tuning.
\item
  \textbf{Metrics and Visualization}: Includes functions for computing
  orthogonality defects (e.g., invariant\_orthogonality\_defect and
  defect\_fast) and plotting optimization traces.
\item
  \textbf{Validation}: Tested on real-world datasets, including the
  Golub leukemia gene expression dataset and the Alzheimer's Disease
  Neuroimaging Initiative (ADNI) dataset, demonstrating maintained or
  improved performance with simplified representations.
\item
  \textbf{Extensibility}: Offers an autograd-compatible variant for
  end-to-end differentiable pipelines and a prototype layer for deep
  learning integration (minimally tested).
\end{itemize}

\subsection{Installation and
Dependencies}\label{installation-and-dependencies}

NSA-Flow can be installed from PyPI using
\texttt{pip\ install\ nsa-flow} or directly from the GitHub repository
via \texttt{pip\ install\ git+https://github.com/stnava/nsa\_flow.git}.
It requires Python ≥ 3.9, PyTorch ≥ 2.0, NumPy ≥ 1.23, and Matplotlib
for visualization. No additional packages are needed for core
functionality.

\subsection{Core Functionality and High-Level
API}\label{core-functionality-and-high-level-api}

The package exposes several key functions for optimization and analysis:

\begin{itemize}
\item
  \texttt{nsa\_flow.nsa\_flow\_orth()}: Autograd-friendly NSA-Flow
  implementation.
\item
  \texttt{nsa\_flow.nsa\_flow\_retract\_auto()}: Applies manifold
  retraction to enforce constraints during optimization and adjusts
  strategies based on the shape of the input matrix.
\item
  \texttt{nsa\_flow.invariant\_orthogonality\_defect()}: Calculates a
  measure of deviation from orthogonality.
\item
  \texttt{nsa\_flow.get\_torch\_optimizer()}: Configures and returns a
  PyTorch optimizer based on user specifications.
\item
  \texttt{nsa\_flow.estimate\_learning\_rate\_for\_nsa\_flow()}:
  Estimates an appropriate initial learning rate for the optimization
  process. Used internally when user elects a \texttt{lr\_strategy}.
\item
  \texttt{nsa\_flow.get\_lr\_estimation\_strategies()}: Returns the
  possible learning rate strategies for the optimization process.
\item
  \texttt{nsa\_flow.test*} \texttt{demo\_*()}: A suite of demonstration
  functions illustrating various aspects of the NSA-Flow algorithm,
  including optimizer tradeoffs and weight parameter effects. There is a
  testing suite that can be run to validate core functionality and will
  automatically call these functions.
\item
  \texttt{nsa\_flow.plot\_nsa\_trace()}: Visualizes the optimization
  trace, showing fidelity and orthogonality metrics over iterations.
  Results from this function are shown in Figure 7.
\end{itemize}

A typical usage example involves initializing a random matrix, running
the optimizer, and evaluating the results:

\begin{Shaded}
\begin{Highlighting}[]
\ImportTok{import}\NormalTok{ torch}
\ImportTok{import}\NormalTok{ nsa\_flow}

\NormalTok{torch.manual\_seed(}\DecValTok{42}\NormalTok{)}
\NormalTok{Y }\OperatorTok{=}\NormalTok{ torch.randn(}\DecValTok{120}\NormalTok{, }\DecValTok{200}\NormalTok{) }\OperatorTok{+} \DecValTok{1}
\BuiltInTok{print}\NormalTok{(}\StringTok{"Initial orthogonality defect:"}\NormalTok{, nsa\_flow.invariant\_orthogonality\_defect(Y))}

\NormalTok{result }\OperatorTok{=}\NormalTok{ nsa\_flow.nsa\_flow\_orth(}
\NormalTok{    Y,}
\NormalTok{    w}\OperatorTok{=}\FloatTok{0.5}\NormalTok{,}
\NormalTok{    optimizer}\OperatorTok{=}\StringTok{"asgd"}\NormalTok{,}
\NormalTok{    max\_iter}\OperatorTok{=}\DecValTok{5000}\NormalTok{,}
\NormalTok{    record\_every}\OperatorTok{=}\DecValTok{1}\NormalTok{,}
\NormalTok{    tol}\OperatorTok{=}\FloatTok{1e{-}8}\NormalTok{,}
\NormalTok{    initial\_learning\_rate}\OperatorTok{=}\VariableTok{None}\NormalTok{,}
\NormalTok{    lr\_strategy}\OperatorTok{=}\StringTok{\textquotesingle{}bayes\textquotesingle{}}\NormalTok{,}
\NormalTok{    warmup\_iters}\OperatorTok{=}\DecValTok{10}\NormalTok{, }\CommentTok{\# standardizes relative weights of fidelity and orthogonality terms}
\NormalTok{    verbose}\OperatorTok{=}\VariableTok{False}\NormalTok{,}
\NormalTok{)}
\NormalTok{nsa\_flow.plot\_nsa\_trace(result[}\StringTok{\textquotesingle{}traces\textquotesingle{}}\NormalTok{])}
\BuiltInTok{print}\NormalTok{(}\StringTok{"Final orthogonality defect:"}\NormalTok{, nsa\_flow.invariant\_orthogonality\_defect(result[}\StringTok{"Y"}\NormalTok{]))}
\end{Highlighting}
\end{Shaded}

This demonstrates the package's ability to refine matrices toward
orthogonality while imposing non-negativity.

\subsection{Specific Demonstrations and
Testing}\label{specific-demonstrations-and-testing}

\subsubsection{nsa\_flow.demo\_nsa\_flow\_optimizer\_tradeoff}\label{nsa_flow.demo_nsa_flow_optimizer_tradeoff}

This demonstration function showcases tradeoffs in optimizer choices
within the NSA-Flow framework. It explores how different optimizers
(e.g., ASGD vs.~others) affect convergence speed, stability, and the
balance between fidelity and orthogonality across different \(w\) with
default settings. An example result is in Figure 4. The example can be
configured with different parameters passed to the function call.

\subsubsection{nsa\_flow.demo\_nsa\_flow\_tradeoff}\label{nsa_flow.demo_nsa_flow_tradeoff}

This function demonstrates general tradeoffs in the NSA-Flow algorithm,
particularly the interplay between the weight \texttt{w} and the
objective values. The result is shown in Figure 5. The example can be
configured with different parameters passed to the function call.

\subsubsection{Automatic Testing}\label{automatic-testing}

The package includes an automatic testing suite that can be executed
from the root directory using
\texttt{python3\ tests/run\_nsa\_flow\_tests.py}. This script runs a
series of unit tests and examples (with visualizations) to validate core
functions, including optimization loops, retraction operators, defect
metrics and deep learning wrappers. It covers prototype features like
the NSA-Flow layer for deep learning (e.g., via
\texttt{tests/test\_nsaf\_layer.py}). Running these tests ensures the
package's stability and correctness across different configurations and
datasets.

\subsection{License and Citation}\label{license-and-citation}

NSA-Flow is released under the MIT License. For research use, cite as:
Stnava et al.~(2025).
\href{https://github.com/stnava/nsa_flow}{NSA-Flow: Non-negative Stiefel
Approximating Flow for Interpretable Representation Learning}.

\section{Session Information}\label{session-information}

\begin{verbatim}
## R version 4.4.2 (2024-10-31)
## Platform: aarch64-apple-darwin20
## Running under: macOS 26.0.1
## 
## Matrix products: default
## BLAS:   /Library/Frameworks/R.framework/Versions/4.4-arm64/Resources/lib/libRblas.0.dylib 
## LAPACK: /Library/Frameworks/R.framework/Versions/4.4-arm64/Resources/lib/libRlapack.dylib;  LAPACK version 3.12.0
## 
## locale:
## [1] en_US.UTF-8/en_US.UTF-8/en_US.UTF-8/C/en_US.UTF-8/en_US.UTF-8
## 
## time zone: America/New_York
## tzcode source: internal
## 
## attached base packages:
## [1] stats     graphics  grDevices utils     datasets  methods   base     
## 
## other attached packages:
##  [1] ggradar_0.2          nnet_7.3-20          randomForest_4.7-1.2 pROC_1.19.0.1        broom_1.0.10        
##  [6] readr_2.1.5          tibble_3.3.0         ggpubr_0.6.2         igraph_2.2.1         fmsb_0.7.6          
## [11] NMF_0.28             cluster_2.1.8.1      rngtools_1.5.2       registry_0.5-1       Rtsne_0.17          
## [16] caret_7.0-1          lattice_0.22-7       golubEsets_1.48.0    Biobase_2.66.0       BiocGenerics_0.52.0 
## [21] MASS_7.3-65          patchwork_1.3.2      gt_1.1.0             ANTsR_0.6.6          DiagrammeR_1.0.11   
## [26] knitr_1.50           scales_1.4.0         tidyr_1.3.1          dplyr_1.1.4          RColorBrewer_1.1-3  
## [31] pheatmap_1.0.13      reshape2_1.4.4       gridExtra_2.3        ggplot2_4.0.0       
## 
## loaded via a namespace (and not attached):
##   [1] IRanges_2.40.1          gaston_1.6              vroom_1.6.6             Biostrings_2.74.1      
##   [5] TH.data_1.1-4           vctrs_0.6.5             effectsize_1.0.1        digest_0.6.37          
##   [9] png_0.1-8               proxy_0.4-27            ppcor_1.1               correlation_0.8.8      
##  [13] bayestestR_0.17.0       parallelly_1.45.1       httpuv_1.6.16           foreach_1.5.2          
##  [17] withr_3.0.2             pgenlibr_0.5.3          xfun_0.54               survival_3.8-3         
##  [21] subtyper_1.3.0          memoise_2.0.1           commonmark_2.0.0        ggbeeswarm_0.7.2       
##  [25] diptest_0.77-2          emmeans_2.0.0           parameters_0.28.2       gmp_0.7-5              
##  [29] visreg_2.8.0            zoo_1.8-14              DEoptimR_1.1-4          Formula_1.2-5          
##  [33] prabclus_2.3-4          rematch2_2.1.2          datawizard_1.3.0        KEGGREST_1.46.0        
##  [37] promises_1.5.0          otel_0.2.0              Evacluster_0.1.0        httr_1.4.7             
##  [41] rstatix_0.7.3           globals_0.18.0          ps_1.9.1                fpc_2.2-13             
##  [45] rstudioapi_0.17.1       UCSC.utils_1.2.0        generics_0.1.4          base64enc_0.1-3        
##  [49] processx_3.8.6          S4Vectors_0.44.0        zlibbioc_1.52.0         lgr_0.5.0              
##  [53] GenomeInfoDbData_1.2.13 xtable_1.8-4            stringr_1.5.2           doParallel_1.0.17      
##  [57] evaluate_1.0.5          hms_1.1.4               colorspace_2.1-2        visNetwork_2.1.4       
##  [61] reticulate_1.44.0       flexmix_2.3-20          magrittr_2.0.4          later_1.4.4            
##  [65] modeltools_0.2-24       palmerpenguins_0.1.1    future.apply_1.20.0     genefilter_1.88.0      
##  [69] robustbase_0.99-6       XML_3.99-0.19           cowplot_1.2.0           matrixStats_1.5.0      
##  [73] class_7.3-23            Hmisc_5.2-4             pillar_1.11.1           nlme_3.1-168           
##  [77] iterators_1.0.14        gridBase_0.4-7          compiler_4.4.2          paradox_1.0.1          
##  [81] stringi_1.8.7           wesanderson_0.3.7       gower_1.0.2             minqa_1.2.8            
##  [85] lubridate_1.9.4         plyr_1.8.9              crayon_1.5.3            abind_1.4-8            
##  [89] locfit_1.5-9.12         bit_4.6.0               sandwich_3.1-1          codetools_0.2-20       
##  [93] multcomp_1.4-29         recipes_1.3.1           paletteer_1.6.0         e1071_1.7-16           
##  [97] plotly_4.11.0           mime_0.13               splines_4.4.2           markdown_2.0           
## [101] Rcpp_1.1.0              VarSelLCM_2.1.3.2       flexclust_1.5.0         smotefamily_1.4.0      
## [105] blob_1.2.4              clue_0.3-66             here_1.0.2              lme4_1.1-37            
## [109] fs_1.6.6                listenv_0.10.0          checkmate_2.3.3         Rdpack_2.6.4           
## [113] ggsignif_0.6.4          estimability_1.5.1      ANTsRCore_0.8.1         coca_1.1.0             
## [117] Matrix_1.7-4            callr_3.7.6             statmod_1.5.1           tzdb_0.5.0             
## [121] pkgconfig_2.0.3         tools_4.4.2             cachem_1.1.0            rbibutils_2.3          
## [125] RSQLite_2.4.3           numDeriv_2016.8-1.1     globaltest_5.60.0       viridisLite_0.4.2      
## [129] DBI_1.2.3               fastmap_1.2.0           rmarkdown_2.30          grid_4.4.2             
## [133] imbalance_1.0.2.1       gprofiler2_0.2.3        sass_0.4.10             coda_0.19-4.1          
## [137] FNN_1.1.4.1             BiocManager_1.30.26     insight_1.4.2           carData_3.0-5          
## [141] rpart_4.1.24            farver_2.1.2            reformulas_0.4.2        mgcv_1.9-3             
## [145] yaml_2.3.10             MatrixGenerics_1.18.1   foreign_0.8-90          ggthemes_5.1.0         
## [149] cli_3.6.5               purrr_1.1.0             stats4_4.4.2            webshot_0.5.5          
## [153] dbscan_1.2.3            lifecycle_1.0.4         mvtnorm_1.3-3           lava_1.8.2             
## [157] kernlab_0.9-33          backports_1.5.0         BiocParallel_1.40.2     annotate_1.84.0        
## [161] timechange_0.3.0        gtable_0.3.6            parallel_4.4.2          limma_3.62.2           
## [165] mlr3cluster_0.1.11      jsonlite_2.0.0          edgeR_4.4.2             bit64_4.6.0-1          
## [169] glasso_1.11             litedown_0.8            RcppParallel_5.1.11-1   dCUR_1.0.2             
## [173] ClusterR_1.3.5          zeallot_0.2.0           timeDate_4051.111       lazyeval_0.2.2         
## [177] shiny_1.11.1            htmltools_0.5.8.1       nmfbin_0.2.1            mlr3pipelines_0.9.0    
## [181] tinytex_0.57            glue_1.8.0              ggstatsplot_0.13.3      XVector_0.46.0         
## [185] rprojroot_2.1.1         mclust_6.1.2            gtsummary_2.4.0         DDoutlier_0.1.0        
## [189] boot_1.3-32             mlr3_1.2.0              R6_2.6.1                broom.mixed_0.2.9.6    
## [193] sva_3.54.0              arm_1.14-4              forcats_1.0.1           labeling_0.4.3         
## [197] GenomeInfoDb_1.42.3     ipred_0.9-15            mlr3misc_0.19.0         nloptr_2.2.1           
## [201] rstantools_2.5.0        tidyselect_1.2.1        vipor_0.4.7             optmatch_0.10.8        
## [205] htmlTable_2.4.3         xml2_1.4.1              car_3.1-3               AnnotationDbi_1.68.0   
## [209] future_1.67.0           ModelMetrics_1.2.2.2    fastICA_1.2-7           statsExpressions_1.7.1 
## [213] rsvd_1.0.5              ciTools_0.6.1           S7_0.2.0                furrr_0.3.1            
## [217] data.table_1.17.8       htmlwidgets_1.6.4       rlang_1.1.6             uuid_1.2-1             
## [221] lmerTest_3.1-3          hardhat_1.4.2           beeswarm_0.4.0          prodlim_2025.04.28
\end{verbatim}

\section*{References}\label{references}
\addcontentsline{toc}{section}{References}

\phantomsection\label{refs}

\end{document}